\documentclass[journal]{IEEEtran}
\IEEEoverridecommandlockouts

\usepackage{graphicx}
\usepackage{array}
\usepackage{cite}
\usepackage[hidelinks]{hyperref}
\usepackage{amsmath}
\usepackage{amsfonts}
\usepackage{amssymb}
\usepackage{hyphenat}
\usepackage{makecell}
\usepackage{orcidlink}
\usepackage{booktabs}
\usepackage{etoolbox}
\usepackage{float}
\usepackage{placeins}
\usepackage{multirow}
\usepackage{stfloats}
\usepackage{enumitem}

\usepackage[caption=false,font=footnotesize]{subfig}
\usepackage{adjustbox}
\usepackage{placeins}
\usepackage{afterpage}
\usepackage{caption}
\usepackage[table]{xcolor}

\newcolumntype{L}[1]{>{\raggedright\arraybackslash}p{#1}}
\newcolumntype{R}[1]{>{\raggedleft\arraybackslash}p{#1}}
\newcolumntype{C}[1]{>{\centering\arraybackslash}p{#1}}

\usepackage{tabularx}
\newcolumntype{Y}{>{\centering\arraybackslash}X}
\newcolumntype{Z}{>{\centering\arraybackslash}p{0.7cm}}

\newcommand{\new}[1]{{\color{black} #1}}

\begin{document}

\bstctlcite{IEEEexample:BSTcontrol}

\title{RF-Informed Graph Neural Networks for Accurate and Data-Efficient Circuit Performance Prediction \\
}

\author{
Anahita~Asadi~\orcidlink{0009-0007-6133-9998},~\IEEEmembership{Graduate Student Member,~IEEE},~
Leonid~Popryho~\orcidlink{0009-0002-0578-9592},~\IEEEmembership{Graduate Student Member,~IEEE},~
and~Inna~Partin-Vaisband~\orcidlink{0000-0002-6399-6672},~\IEEEmembership{Senior Member,~IEEE}
\thanks{This work was supported in part by the CogniSense: Center on Cognitive Multi-spectral Sensors, one of seven centers in Joint University Microelectronics Program (JUMP) 2.0, a Semiconductor Research Corporation (SRC) program sponsored by the Defense Advance Research Project Agency (DARPA).}
\thanks{A. Asadi, L. Popryho, and I. Partin-Vaisband are with the Department of Electrical and Computer Engineering, University of Illinois Chicago, Chicago, IL 60607 USA (e-mail: aasadi5@uic.edu; lpopry2@uic.edu; vaisband@uic.edu).}
}

\maketitle

\begin{abstract}

Accurately predicting the performance of active radio frequency (RF) circuits is essential for modern wireless systems but remains challenging due to highly nonlinear
behavior and the high computational cost of traditional simulation tools. 
Existing machine learning (ML) surrogates often require large datasets to generalize across various topologies or are not accurate on held-out circuits. 
This work presents a lightweight, data-efficient, and topology-aware graph neural network (GNN) framework for predicting key performance metrics of active RF circuit classes, such as low-noise amplifiers (LNAs), mixers, voltage-controlled oscillators (VCOs), power amplifiers (PAs), and voltage amplifiers (VAs).
The proposed framework employs RFIC domain-informed feature indexing to enable cross-topology adaptability by cheap encoding of functional device semantics (e.g., differential pair and varactor transistors) and efficient knowledge transfer. The surrogate model represents circuits using device-terminal graph abstractions to preserve fine-grained connectivity and transistor-level symmetry. The final model is generalized to a wide variety of classes by being trained in parallel.
Experimental results demonstrate accurate modeling of multimodal and heavy-tailed RF performance distributions, achieving an average mean relative error (MRE) of 2.71\% across nineteen topologies, an improvement of 3.3$\times$ and 20$\times$ faster in training over prior art, and the generalization to held-out topologies is improved by $\sim$26.2$\times$. Furthermore, this work shows $\sim$36$\times$ training data efficiency compared to state-of-the-art, demonstrating its effectiveness for scalable and deployment-ready RF design automation.
\end{abstract}


\section{Introduction}
\label{sec:I}

With the growing importance of modern wireless systems (e.g., the Internet of Things \cite{b1}, 5G \cite{b2} RADAR \cite{b3}, and LiDAR \cite{b4}), accurate modeling and optimization of RF integrated circuits (RFICs) is more critical than ever. The performance of key building blocks of such systems, ranging from power amplifiers (PA) to transmitters, directly affects fidelity, efficiency, and robustness of modern systems. However, they operate in highly nonlinear and layout-sensitive regimes, making accurate modeling particularly challenging.

While highly accurate, traditional simulators (e.g., SPICE, ANSYS) are computationally expensive
\new{for the extensive design-space exploration required in RFIC design, relying on costly numerical methods (e.g., trapezoidal, Gear, and Newton-Raphson integration) and, for electromagnetic analysis, finite element method (FEM) solvers \cite{b5}.}
\new{Furthermore, they both scale poorly with advancing technologies, such as heterogeneous integration and packaging effects \cite{b6}.}
\new{Consequently, the substantial expert effort needed for topology selection and device sizing highlights the need for accurate and fast automation methods.}

\new{A broad range of automation strategies has been explored to alleviate this cost, from heuristic \cite{b25} and meta-heuristic optimizers \cite{b8, b9, b10}}
\new{to classical \cite{b7},}
\new{machine learning (ML) \cite{r1},}
\new{and neural surrogate models \cite{b11, b14, b15, b16, b24, b26, b28, r5, b27} that predict circuit figures of merit (FoMs) directly from design parameters.}
\new{Despite steady progress, these approaches typically require large simulated datasets and substantial compute, and they generalize poorly across circuit topologies, demanding exhaustive retraining with structural changes. The issue stems from most surrogates not explicitly representing circuit topology.}

\new{Graph neural networks (GNNs) address this gap directly by naturally encoding circuit connectivity, reducing or eliminating retraining when adapting to new designs. }
\new{However, existing GNN frameworks are largely tailored to analog blocks and do not generalize across the broad range of RFIC classes and performance metrics \cite{g1, g5, g7, g8, g9, g10, g11}.}
\new{However, in practice, RFIC classes (e.g., VCO, LNA) are selected first, and only then explored across topologies and sizing values. Therefore, such unified cross-class models offer limited practical benefit relative to their training overhead.}

\new{While the aforementioned approaches offer valuable design automation capabilities,}
\new{this work offers a data-efficient and computationally-feasible framework.}
\new{To our knowledge, this is the first GNN model that accurately scales across a variety of RFIC classes and generalizes well to unseen circuit topologies using an RF-informed inductive bias, enabled through the class-specific model.}
The main contributions are:
\begin{itemize}
    \item \new{\textbf{Conceptual approach for class-specific RF surrogate modeling.} 
    Restricting training to a single RFIC class reduces cross-topology diversity, degrading unified model accuracy by $\sim$2.3$\times$ under topology-specific evaluation and $\sim$1.2$\times$ in cross-topology transfer. Class-specific modeling framework is introduced that aligns surrogate modeling with practical RFIC design workflows while reducing training overhead by 20$\times$ compared to unified surrogate models.}

    \item \textbf{RF-informed GNN architecture.} Device functionality is embedded directly into the graph representation through an RF-aligned feature indexing. Unlike prior approaches that rely on auxiliary encoder networks, teacher models, or large datasets to infer device roles implicitly, the proposed representation enables the GNN to learn transferable electrical interactions directly. Compared to prior art, this approach decreases the average weighted mean relative error (MRE) across nineteen topologies to 2.71\%, an improvement of $\sim$3.3$\times$.

    \item \new{\textbf{Cross-topology transfer by fine-tuning a single GNN layer.} The RF-informed indexing concentrates topology-specific information in the first layer, so adapting it alone suffices. This framework achieves an average of $\sim$1.1\% MRE on held-out topologies through only fine-tuning the first GNN layer and over-performs the prior art by $\sim$26.2$\times$.}

    \item \textbf{Data-efficient, fast-inference RF surrogate model.} The proposed approach reduces the SPICE training data required for relative to prior state-of-the-art methods, \new{achieving a sample-efficiency improvement of $\sim$36$\times$.} Compared to Spectre, the proposed surrogate provides inference speedups of $5.6\times10^3$ on CPU and $4.2\times10^4$ GPU.
\end{itemize}

\new{While the primary objective of this work is schematic-level RFIC performance prediction,
the proposed surrogate can directly replace the schematic-level performance predictor in existing post-layout optimization frameworks (e.g., FALCON \cite{g19}) while leaving the remaining optimization pipeline unchanged.} 

The remainder of this paper is organized as follows. Recent advances in \new{non-GNN} and GNN-assisted RF circuit modeling are reviewed in Section~\ref{sec:II}.
The proposed GNN-based framework, including graph representations and model architecture, is detailed in Section~\ref{sec:III}. The experimental setup is described in Section~\ref{sec:IV}, and results with comparative analyses are presented in Section~\ref{sec:V}. Conclusions and directions for future work are provided in Section~\ref{sec:VI}.

\begin{table*}[b]
\centering
\caption{Comparison of State-of-the-Art ML Frameworks.}
\small

    \begin{tabular}
    {C{0.8cm}C{2.6cm}C{2.4cm}C{2.9cm}C{2.0cm}C{4.9cm}}
    \toprule
    \textbf{Work} & \textbf{Key Architectural Features} & \textbf{Target\hspace{20pt} Circuits} &  \textbf{Error\,per Dataset\,Size} & \textbf{Circuit Limitations} & \textbf{Other\hspace{100pt} Limitations} \\
    \hline
    \makecell[tc]{\textbf{This}\\\textbf{work}} & GNN & LNA, Mixer, VCO, PA, VA & \makecell[tc]{MRE\textsubscript{$\sim$18.5k}$\approx$9.09\% \\ MRE\textsubscript{$\sim$665k}$\approx$2.71\%} \\
    \hline
    \cite{g19} & GNN & LNA, Mixer, VCO, PA, VA & MRE\textsubscript{$\sim$665k}$\approx$9.09\% & --- & \makecell[tl]{Prohibitive data size; Struggles with\\cross-topology generalization; Overhead\\of an Extra Encoder before GNN} \\ 
    \hline
    \cite{g8} & GNN + TL & Resistor ladders, VA & \makecell[tl]{Acc@200\textsubscript{20k+126k}$\approx$9\%} & \makecell[tc]{simple \\ analog} & \makecell[tl]{Mediocre accuracy in heterogeneous\\RF circuit training; Struggles with cross\\-topology generalization; DC simulation\\required for fine-tuning} \\
    \hline
    \cite{g14} & GNN + Text Embedding + Attn & VA & MRE\textsubscript{1,490+447}$\approx$11.8\% & \makecell[tc]{simple \\ analog} & \makecell[tl]{Mediocre accuracy; Inaccurate for\\complex RF metrics; Overhead\\of Extra Encoders} \\
    \hline
    \cite{b17} & CNN + LSTM & two-port RF filters & MSE\textsubscript{8k}$\approx$0.0015 & passive & \makecell[tl]{Struggles with cross-topology\\generalization} \\
    \hline
    \cite{b18} & Convolutional AE + Regressor & Band-pass filter & MSE\textsubscript{5,929+916}$\approx$0.001 & passive & \makecell[tl]{Struggles with cross-topology\\generalization} \\
    \hline
    \cite{b16} & CNN + GA & PA matching networks & RMS\textsubscript{500k+170k}$\approx$0.4 & passive & \makecell[tl]{Prohibitive data size; Struggles with\\cross-topology generalization} \\
    \hline
    \cite{b22} & DeepGRU + Dropout & VA, VCDL, DLL & --- & analog & \makecell[tl]{Inaccurate for complex RF metrics;\\Struggles with cross-topology\\generalization} \\
    \hline
    \cite{b23} & DeepRNN & High-speed interconnects & Error\textsubscript{45~WFs$\times$300~steps} $\approx$0.0018 & passive & \makecell[tl]{Struggles with cross-topology\\generalization} \\
    \hline
    \cite{b24} & Topology-aware S-TCNN & High-speed digital interconnects & NMSE\textsubscript{$<$750}$\approx$2.17 & passive & \makecell[tl]{Struggles with cross-topology\\generalization} \\
    \bottomrule
    
    
    \end{tabular}
\label{tab:ml_comparison}
\end{table*} 

\section{Related Work} \label{sec:II}

\new{Automation strategies for RFIC design span from classical search and optimization to learning-based surrogate models. Both directions are surveyed in this section. Non-GNN methods (i.e., heuristic and meta-heuristic optimizers and conventional ML-based surrogates) are first reviewed for their scalability and generalization limitations in high-dimensional, nonlinear RF design spaces. GNN-based approaches are then reviewed as a more generalizable alternative. The remaining gaps in accuracy, data efficiency, and cross-class generalization are highlighted to motivate the proposed framework.}

\subsection{Non-GNN Methods}
\label{subsec:nonGNNhistory}

A broad spectrum of heuristic methods has been explored in the area of RFIC optimization. Simple local-search heuristics, such as direct binary search (DBS), have been used in \cite{b25} for a dual-band filter. However, such greedy approaches are not scalable and are prone to getting trapped in FoM local optima. To escape this issue, meta-heuristic optimizers such as evolutionary algorithms (EAs) \cite{b8,b9} and Bayesian optimization (BO) \cite{b10} have been proposed. In higher dimensional RF problems, however, EAs become sample-inefficient and exhibit poor reproducibility. BO also struggles to scale, as multi-dimensional Bayesian acquisition function maximization leads to long run-times and premature convergence. 

An alternative strategy is to learn a surrogate model for fast prediction of circuit FoMs and integrate it into the optimization process. Classical deterministic surrogates, such as partial least-squares (PLS) regression \cite{b7}, offer weighted combination of all inputs that perform well when the response surface is nearly linear but do not capture the strong nonlinearities and complex FoM distributions inherent in RFICs.
Traditional ML regressors, such as k-nearest neighbors (kNNs), random forests, and support vector regressors (SVRs), have been used for inferring circuit parameters that produce a desired performance (i.e., inverse design), achieving mean relative errors (MRE) as low as 0.23\% for receiver components consisting of voltage amplifiers (VA), LNAs, and mixers, at a single frequency \cite{r1}. However, errors rise to $\sim$50\% in nonlinear circuits such as power amplifiers (PAs). In contrast, deep neural surrogates can capture complex nonlinear interactions better, making them more suitable for generalizable modeling and downstream optimization.

A growing body of work has applied neural networks (NNs) in RF domains to model various FoMs from sizing parameters or perform inverse design \cite{b11}. 
One of the simplest NN architectures, feed-forward neural networks have been widely used for modeling passive RF circuits \cite{b14, b15}. Alternatively, for more complex circuits with nonlinear active components, such as PAs, both local and global interactions play an important role in circuit behavior, making convolution neural networks (CNNs) an intuitive choice.
As an example, in \cite{b16}, the insertion loss S-parameters of a post-layout PA matching circuit is modeled with a CNN and optimized with genetic algorithm (GA). This approach achieves approximately 4,500$\times$ speedup in the inverse design process. 
However, this models generalize only to circuits within the same distribution the CNN was trained on. New circuits exhibiting significantly different physics, such as ones with active components, often require exhaustive retraining, potentially offsetting the benefits of automation. Moreover, this work requires a large dataset of over 300{,}000 simulations and substantial computational and memory resources.

NN-based approaches have also been enforced with physics rules to satisfy fundamental laws, such as causality and passivity in \cite{b24}. This work uses two custom NN layers, one layer enforces consistency with the Kramers-Kronig relations (fundamental causal constraints that link the real and imaginary parts of the frequency response) via Hilbert transform. The other layer enforces passivity using a differentiable minimum-phase filter constructed during training. Another work, \cite{b26}, uses a causality-aware UNet CNN surrogate trained on $\sim$4,000 Ansys HFSS simulations and combined with BO to optimize an impedance-transforming network.
However, incorporating device functionalities in active RFICs into ML models for generalization across different topologies remains an open question.

Reinforcement learning (RL) has also been explored for PA circuit topology and sizing optimization in \cite{b28}. However, RL methods typically require extensive simulations, making them less suitable for large-scale systems. Generation of the 35{,}000 training dataset in \cite{b28} requires $\sim$3.3 days on a 192-core cluster. Furthermore, RL-based approaches require data-intensive retraining and substantial adaptation to generalize across circuit classes with substantially different behavior or FoM distribution \cite{r5}.

Some prior work tried to circumvent topology-awareness by characterizing circuits through their building blocks. 
The subcircuits of RF circuits are abstracted by two-port S-parameters in \cite{b27}, where an NN infers the behavior of larger topologies from these subblocks, achieving an average $R^2$ of 95\% on phase shifters and LNAs. A related approach \cite{g21} instead combines subcircuits using their block-level specifications (gain, power, input/output impedance) to model OpAmps and filters, reaching an average MAPE of $\sim$0.06 on filters.
However, such behavioral interfaces are quasi-linear abstractions, as fundamental-frequency S-parameters and cascaded block specifications cannot capture the behavior of highly nonlinear circuits such as mixers and PAs. 
Moreover, they both remain implicitly tied to a fixed structure (e.g., the number, sequence, and port count of sub-blocks), changing which requires retraining of the network. \new{Data inefficiency, sensitivity to topology changes, and the absence of an explicit structural representation, motivate models that naturally encode circuit topology, which is discussed next.}

\begin{figure*}[htbp]
  \centering
  \includegraphics[width=1\textwidth]{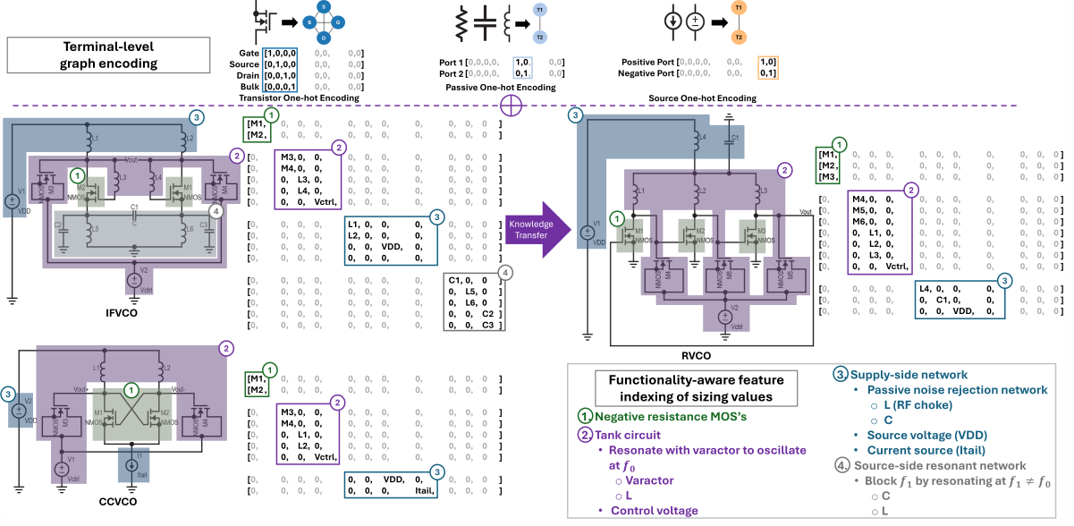}
  \caption{Proposed circuit-to-graph conversion and feature extraction with functionality-aware feature indexing. \new{Top: terminal-aware encoding common to all circuits, distinguishing transistor terminals (e.g., gate and source), passive-element terminals, and source terminals (positive and negative). Bottom: functionally-aware encoding, where circuit elements are indexed according to the RF subcircuit to which they belong (e.g., negative-resistance MOS pair, tank circuit, supply-side network, or source-side resonant network in VCOs). Since these functional subcircuits are standard RFIC building blocks, the required one-time expert labeling is inexpensive and reproducible.}}
  \label{fig:Circuit2Graph}
\end{figure*}

\subsection{GNN-Based Approaches}
\label{subsec:GNNhistory}

GNNs are a class of deep learning models designed to operate on graph-structured data by learning node, edge, or graph-level representations via message passing mechanisms. In their earliest form, recurrent GNNs propagate information iteratively across the graph until convergence, but incur high computational and memory cost. Algorithms such as gated GNNs were developed to address scalability issues by using GRUs to limit message propagation to a fixed number of steps \cite{g0}. GNNs in circuits were first introduced to model S-parameters of filters with significant accuracy degradation in unseen circuits \cite{g1}. 

Spatial GNN layers, such as graph isomorphism networks, are used in \cite{g9} to capture performance of some topologies of operational-amplifiers (OpAmps) and transfer them to held-out ones in zero- and few-shot settings. Other spatial GNN layers, such as graph attention network (GAT), offer lower computational cost and better accuracy for digital circuits (e.g., in evaluating transistor degradation and threshold voltage shifts in digital circuits \cite{g4}). However, they are found to be less competitive than graph convolutional network (GCN) layers for analog and RF circuit modeling. For example, \cite{g20} applies GCNs for circuit identification and \cite{g5} combines GCN with RL for transistor sizing optimization. Transfer learning is used to generalize GCN knowledge in OpAmps across different technologies \cite{g5, g10}, process, voltage, and temperature (PVT) variations \cite{g11}, and circuit topologies \cite{g9}, albeit with some trade-off in accuracy.

However, as the network depth increases, GCNs tend to suffer from vanishing gradients, over-smoothing, and overfitting, leading to degraded performance. To stabilize the learning process of deep GCNs, pre-activation of residual connections and message normalization are introduced in generalized graph convolution (GENConv) layers \cite{g7}. In \cite{g8}, GENConv is used to predict DC voltage of each node in resistor ladders and OpAmps from sizing values, and later fine-tuned to predict AC output voltage and gain. 
However, the current GNN approaches either do not generalize accurately and data-efficiently to a broad range of unseen RF circuits or require a separate computationally expensive encoder.

GNNs are also used to predict post-layout effects in passive or analog circuits. Coupling capacitance and parasitic resistance are predicted in \cite{g17} with a relative error of 18.6\%. Chip-scale interconnect capacitance has been extracted in \cite{g16} with an absolute relative error less than 10\% on average.
In \cite{g14}, a self-supervised GNN is combined with text and self-attention models to predict distance between devices. This latent representation is then transferred to predict wire length with $R^2$ of $\sim$57\% and parasitic-capacitance tasks with relative error of 11.8\%. 

However, applying existing GNNs to active RF circuits presents several challenges. RF circuits show strong nonlinear behavior and tight component dependencies (e.g., matched pairs, feedback loops) which produce high-tailed, skewed and multimodal distributions of FoMs, which are difficult to model with typical methods. To address this, a logarithmic normalization is applied to GNNs in \cite{g18} to stabilize variance and improve learning. The relative error, however, remains as high as 20\% in VCO net-level parasitic capacitances prediction. \new{Nevertheless, all these methods predict layout and parasitic quantities rather than circuit performance metrics.
To obtain the performance of interest, the estimated layout-effects are co-optimized with schematic-level performance \cite{g19}. Therefore, accurate and data-efficient schematic-level performance prediction remains the bottleneck of existing post-layout RFIC optimization frameworks.}

The most representative ML-based techniques, along with their strengths and limitations, are summarized in Table~\ref{tab:ml_comparison}. Based on the qualitative comparison, \cite{g19} is the most promising approach for fast and accurate performance prediction in RFICs. An impressive MRE of 9.09\% in predicting key performance metrics across a variety of RF circuits, including LNAs, VCOs, mixers, and PAs is achieved in \cite{g19}.
However, this performance comes at the cost of training on $\sim$35{,}000 
\footnote{Per-circuit training dataset sizes are not reported in \cite{g19}. To enable fair comparisons, the dataset sizes are estimated and experimentally verified by inspection of the released code, which employs a 70\%/15\%/15\% training/validation/test partition. Out of 1 million total clean dataset samples, an implied dataset size of \new{$\sim$35{,}000} samples per topology is obtained.}
samples per circuit topology, a prohibitively large dataset. 

\section{Proposed Framework}
\label{sec:III}

\begin{figure}[b]
  \centering
  \vspace{-15pt}
  \includegraphics[width=0.45\textwidth]{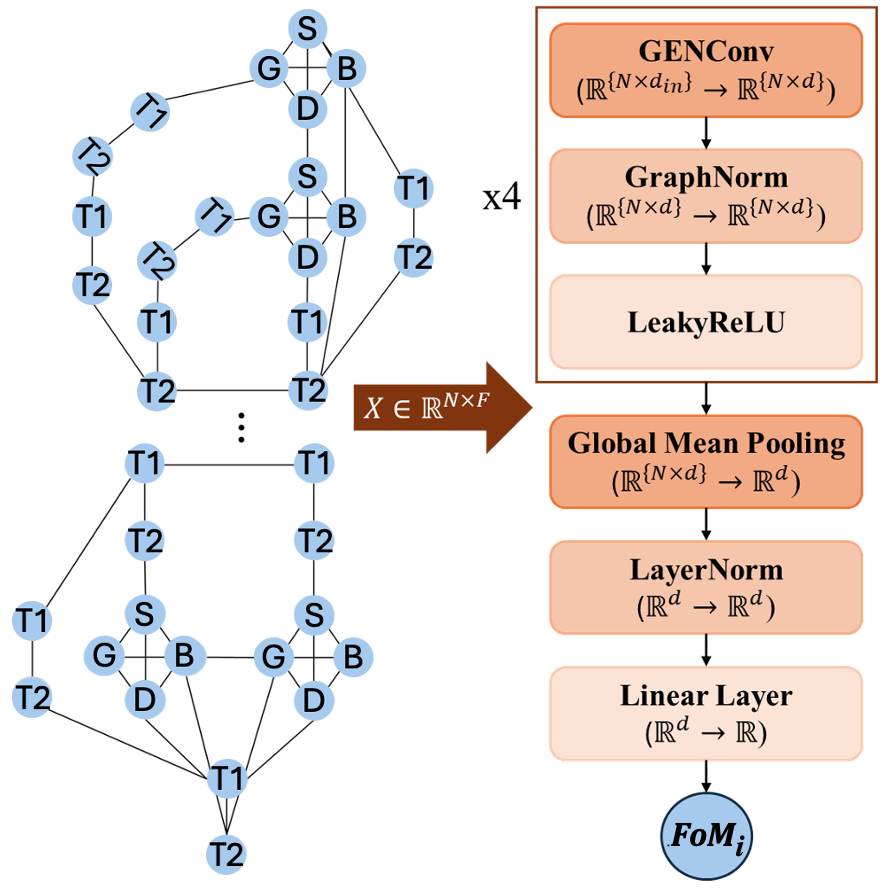}
  \caption{Model Architecture: GENConv layers \new{can be interpreted as analogous to} KVL and KCL, while GraphNorm enables graph structure-awareness and numerical stability.}
  \label{fig:Architecture}
\end{figure}

The proposed end-to-end framework for predicting RF circuit performance metrics from netlist-derived graph representations is described in this section. First, the translation of RF circuits into annotated graphs is formalized to capture both inter-component relationships and global structure. Then, the GNN architecture designed to process these graphs is presented, enabling scalable circuit modeling.

\subsection{Graph Representation of RF Circuits}
\label{subsec:RFInformed}

Each RF circuit is represented as an undirected graph where graph nodes correspond to component terminals, and graph edges encode electrical connectivity. Terminals of the same component (e.g., transistor source, drain, gate, and body bias terminals) form fully-connected subgraphs.
Each graph node (i.e., component terminal) is annotated with a feature vector, encoding physical parameters (e.g., transistor width$\times$fingers and length, power supply levels, resistor, capacitor, inductor impedances).

\new{Circuit components serving different functionalities are assigned unique indices in the feature vector, while functionally similar components are intentionally mapped onto the same indices to preserve role equivalence.
This manual indexing scheme embeds RFIC domain knowledge into the model representation, introducing RF-informed inductive biases that improve intra-class accuracy in Subsection~\ref{subsec:ablation}.
As an example, some VCO topologies are shown in Figure~\ref{fig:Circuit2Graph}. The negative resistance MOS pair (M1, M2 in IFVCO and CCVCO; M1, M2, M3 in RVCO) serves a different purpose than MOS varactors (M3, M4 in IFVCO and CCVCO; M4, M5, M6 in RVCO). This difference is captured by positioning the physical parameters in different feature vector indices. Similarly, tank circuit inductors (L3, L4 in IFVCO; L1, L2 in CCVCO; L1, L2, L3 in RVCO) have the same index position, but different positioning than supply-network RF choke inductors (L1, L2 in IFVCO; L4 in RVCO). This tailored feature engineering imparts circuit design intuition to the model and facilitates cross-topology generalization in Subsection~\ref{subsec:TL}.} 

One-hot encoding is employed to represent terminal position within a transistor (first four positions, e.g., '\underline{1000}0000' for gate, '\underline{0100}0000' for source), passive element (second two positions, e.g., '0000\underline{01}00' and '0000\underline{01}00' for each terminal), and current or voltage source (third two positions, e.g., '000000\underline{10}' for positive and '000000\underline{01}' negative pole), allowing the model to distinguish current flow direction and identify terminal roles. \new{Terminals of a device share the same physical parameters.}

\subsection{GNN Architecture}

\new{The following circuit-theoretic descriptions are offered as intuition to motivate the architectural choices and results in Subsection~\ref{subsec:TL}, not as enforced constraints. Consequently, this work does not claim to include mechanisms imposing physical behavior, and the domain knowledge in this work is introduced explicitly through the RF-informed feature indexing of Subsection~\ref{subsec:RFInformed}, the impact of which  is validated by the ablation studies in Subsection~\ref{subsec:ablation}.}

The architecture of the proposed model is shown in Figure~\ref{fig:Architecture}. 
The netlist-derived graph $G=(V,E)$ is fed to the GNN, which consists of four GENConv layers followed by linear prediction heads for each circuit performance. Nodes carry an initial feature vector $h_v \in \mathbb{R}^d$. An independent GENConv stack and head is instantiated per-output metric or FoM:
\begin{itemize}
    \item \textbf{First layer (1-hop neighborhood)}: A terminal aggregates signals from other terminals of the same component and from directly connected nets.  This supports transistor-level property inference (e.g., drive strength, capacitive loading).
    \item \textbf{Second layer (2-hop neighborhood)}: Through an additional message-passing step, the terminal receives information from adjacent components, enabling the modeling of inter-component interactions in matching networks and load.
    \item \textbf{Third layer (3-hop neighborhood)}: Information propagates across short multi-stage paths or local feedback loops. This captures mini subcircuit effects such as inter-stage loading, peaking networks, and first-order crosstalk.
    \item \textbf{Fourth layer (4-hop neighborhood)}: The receptive field expands to longer functional chains or distributed networks, providing context for full-subcircuit behavior such as biasing or matching networks, long cascade chains, and far-end impedance lines. 
\end{itemize}

Let $h_v^{(\ell,i)}$ denote the hidden representation of node $v$ at layer $\ell$ for output $i$. 
The message passed from node $u$ to node $v$ at layer $\ell$ and output or FoM $i$, a given circuit metric, is

\begin{equation}
    m_{uv}^{(\ell,i)} = \phi_{\text{msg}}^{(\ell,i)}\!\bigl(h_u^{(\ell,i)}\bigr), \qquad \ell = 0, \ldots, L-1
\end{equation}
where $\phi_{\text{msg}}$ is a learnable transformation in an $L$-layer GNN.
The message \new{resembles} an incremental stimulus propagated from terminal represented by $u \in V$ to its adjacent terminal $v \in V$, with $V$ set of vertices. Neighbor contributions are modulated by attention-like coefficients computed via softmax,
\begin{equation}
    \alpha_{uv}^{(\ell,i)} = \frac{\displaystyle\exp\!\Bigl(a^{(\ell,i)}\bigl(h_u^{(\ell,i)},h_v^{(\ell,i)}\bigr)\Bigr)}
         {\displaystyle\sum_{k\in\mathcal N(v)}\exp\!\Bigl(a^{(\ell,i)}\bigl(h_k^{(\ell,i)},h_v^{(\ell,i)}\bigr)\Bigr)},
\end{equation}
where $\mathcal{N}(v)$ is the set of neighboring nodes of $v$, and $a(\cdot,\cdot)$ is a learned scoring function. Intuitively, the attention weight $\alpha_{uv}^{(\ell,i)}$ plays the role of a normalized admittance ${g_{uv}/\sum g_{uv}}$, whereby neighbors with higher effective conductance (and thus, higher current) exert greater influence on the aggregation. The aggregated message at node $v$ is, therefore,
\begin{equation}
    M_v^{(\ell,i)} = 
    \sum_{u\in\mathcal N(v)}\alpha_{uv}^{(\ell,i)}\,m_{uv}^{(\ell,i)}.
\end{equation}

Traditional simulators apply Newton-Raphson iterations to enforce Kirchhoff’s current law (KCL) by iteratively adjusting node voltages to minimize the total current mismatch at each node. The net mismatch (i.e., the residual current that remains after evaluating all branch currents based on the current voltage estimate) indicates how far the solution is from satisfying KCL. 
Message passing in GNNs resembles superposition, where interactions are accumulated from adjacent elements and then transformed by learnable update operators. The coefficient ($\alpha(\cdot,\cdot)$) regulates how strongly one node information ($m$) influences another. $M$ is the mismatch in aggregated messages at a node before being resolved by the model’s update function, similar to the net current mismatch before a Newton-Raphson update. Thus,
the learnable function in the GNN refines node embeddings to encode local electrical interactions and global circuit context.

Formally, the node-wise pre-activation output of GENConv is
\begin{equation}
    z_v^{(\ell,i)} =
    \mathrm{GENConv}^{(\ell,i)}\!\bigl(h^{(\ell,i)}, E\bigr)_v.
\end{equation}
GraphNorm is then applied as in
\begin{equation}
    \tilde z_v^{(\ell,i)} =
    \mathrm{GN}\!\bigl(z_v^{(\ell,i)}\bigr),
\end{equation}
to normalize node features after each message passing. GraphNorm follows a formulation similar to the standard feature scaling, but the normalization statistics (mean and variance) are computed independently for each graph. GraphNorm is beneficial for enabling cross-topology generalization, as it stabilizes internal representations (i.e., message magnitudes) while preserving circuit-level independence. From a physical perspective, GraphNorm can be interpreted as removing a graph-level common-mode component from the learned signals and rescaling their effective gain, similar to bias referencing and automatic gain control in circuit analysis.

The GraphNorm normalization is followed by a nonlinear activation function,
\begin{equation}
    h_v^{(\ell+1,i)} = \operatorname{LeakyReLU}\!\bigl(\tilde z_v^{(\ell,i)}\bigr).
\end{equation}
The \texttt{LeakyReLU} activation is similar to piecewise-linear I-V characteristic: the positive-slope region resembles the forward-conducting path of a device, while the small negative-slope ``leak'' term prevents node isolation.

Stacking $L$ such layers is analogous to cascading $L$ successive linearization-and-update steps, similar to the iterative procedures used in nonlinear AC small-signal solvers. The node-level embeddings produced by the final layer are then aggregated to form a single graph-level embedding,
\begin{equation}
    h_G^{(i)} = 
    \operatorname{MeanPool}\!\bigl\{h_v^{(L,i)}:v\in V\bigr\}
    \;=\;
    \frac{1}{|V|}\sum_{v\in V}h_v^{(L,i)},
\end{equation}
where $\mathrm{MeanPool}$ takes the average of all terminal-level embeddings, effectively representing all FoMs with a single vector $h_G^{(i)}$. Prediction vectors are then normalized through LayerNorm to form $\mathrm{LN}(h_G^{(i)})$, which forms the final output scalar $y^{(i)}$,
\begin{equation}
    \hat y^{(i)} = 
    \mathbf w^{(i)\!\top}\mathrm{LN}\left(h_G^{(i)}\right)+b^{(i)}.
\end{equation}

\section{Experimental Setup}
\label{sec:IV}

All experiments are performed on NVIDIA GeForce RTX 4090 GPUs using floating-point 32 precision. Circuit classes are trained independently but concurrently using GPU-parallel workloads, allowing multiple specialized surrogates to be optimized simultaneously rather than sequentially within a unified model.

The framework employs GENConv and GraphNorm from \texttt{PyTorch Geometric} \cite{r2}.
Reproducibility is ensured by fixing random seeds across all relevant libraries and disabling residual nondeterminism in CUDA Deep NN kernels.
Further details, including the training strategy, loss formulations, data preprocessing and augmentation, and computational settings are presented in this section. The framework is designed to enable efficient training.

\subsection{Dataset Description}
A publicly available \new{schematic simulation data} first introduced in FALCON \cite{g19} is used. Approximately 4.5\% of the dataset, consisting of duplicate, contradictory (i.e., identical sizing values associated with different output metrics), and physically impossible (e.g., zero-value output power or bandwidth) entries are removed. 
All reported sample counts refer to the cleaned dataset. The dataset comprises over one million Cadence-simulated analog and RF circuits of various classes and topologies such as single- and double-balanced active (SBAMixer, DBAMixer) and passive (SBPMixer, DBPMixer) mixers; cross-coupled (CCVCO), Colpitts (ColVCO), inductive-feedback (IFVCO), and ring (RVCO) VCOs; common-source (CSLNA), common-gate (CGLNA), cascode (CLNA), and differential (DLNA) LNAs; as well as class-B (ClassBPA), class-E (ClassEPA), differential (DPA), and Doherty (DohPA) PAs.



Each data point is a with given sizing parameters. Simulation results are stored after parameter sweeps, and span key performance metrics such as power gain (P\textsubscript{Gain}), conversion gain (C\textsubscript{Gain}), bandwidth (BW), noise figure (NF), DC power consumption (P\textsubscript{dc}), drain efficiency (DE), saturation power (P\textsubscript{SAT}), oscillation frequency ($f$\textsubscript{osc}), tuning range (TR), output power (P\textsubscript{out}), phase noise (PN), voltage swing (V\textsubscript{swg}), power-added efficiency (PAE), and S-parameters at the fundamental frequency. 

\subsection{Initialization}

Node features are min-max-scaled while the prediction targets are z-standardized. Statistics are computed on the training subset only and reused for validation and test, eliminating leakage of those sets into train. 

\subsection{Hyperparameters}

Each individual GNN comprises four layers with 128 hidden features. \texttt{LeakyReLU} activations have a slope of 0.1. Graphs are fed to train and test pipelines in batches of 128. The data is augmented with Gaussian noise drawn from $\mathcal{N}(0,0.01)$ to discourage overfitting.
Weighted Adam (\texttt{AdamW}) optimizer is used. 
Learning rate is reduced by a factor of two with the \texttt{ReduceLROnPlateau} scheduler if the median validation loss does not improve for five consecutive epochs. Early stopping is applied individually to each output (i.e., per FoM) head: a given prediction head is frozen if it shows no improvement for ten consecutive epochs, while training continues for the remaining heads. This strategy accommodates varying convergence rates across different performance metrics, preserving model capacity for more challenging targets while avoiding overfitting on simpler ones.

\subsection{Loss Function}

The GNN outputs are trained to minimize \texttt{SmoothL1Loss} after applying min-max normalization and z-score standardization.

\subsection{Computational Cost}

For a batch of graphs with a total of $N_b$ nodes and $E_b$ edges, the computational complexity of the GNN-based framework scales as $\mathcal{O}\!\left(D \left[ E_b L H \;+\; N_b \big(F H + (L-1) H^2\big) \right] \right)$, where $D$ is the number of predicted FoMs (outputs), $L$ is the number of message-passing layers, $F$ is the input feature dimension, and $H$ is the hidden dimensionality. Among the 20 topologies, the average per-topology number of nodes, edges, degree, and feature dimensions are 33, 100, 3, and 23, respectively.

\subsection{Evaluation Metrics}
The framework accuracy is evaluated using standard relative benchmarks: (i) average MRE, 
(ii) symmetric mean absolute percentage error (sMAPE), and (iii) root mean square error (RMSE). The accuracy benchmarks are determined based on the non-scaled (i.e., inverse-transformed), actual target values.
RFIC performance metrics are often multimodal and are therefore rarely well characterized by a single mean value. Therefore, mean-based evaluation benchmarks such as $R^2$ and NRMSE can be misleading and are not reported.

The reported weighted MRE is computed as the sample-count-average weighted of per-metric mean relative error across circuit classes. Specifically, for each metric and class, MRE is first computed as the mean absolute relative error over all valid test samples:
\begin{equation}
    \text{MRE}_{c,k} =
    \frac{1}{N_{c,k}}
    \sum_{i \in \mathcal{V}_{c,k}}
    \left| \frac{\hat{y}_{c,i,k} - y_{c,i,k}}{y_{c,i,k}} \right|,
\label{eq:MRE}
\end{equation}
where $\mathcal{V}_{c,k}$ denotes the set of valid samples and $N_{c,k}$ its size. The final reported weighted MRE is then computed as
\begin{equation}
    \text{MRE}_k^{\mathrm{ave}} =
    \frac{\sum_{c=1}^{C} N_{c,k} \cdot \text{MRE}_{c,k}}
    {\sum_{c=1}^{C} N_{c,k}},
\label{eq:WeightedMRE}
\end{equation}
ensuring that each test sample contributes proportionally to the final error. Samples with non-finite targets or $|y_{c,i,k}| \le 10^{-6}$ are excluded to ensure numerical stability.
Similarly, the sMAPE is computed as
\begin{equation}
    \text{sMAPE}_{c,k} =
    \frac{1}{N_{c,k}}
    \sum_{i \in \mathcal{V}_{c,k}}
    \frac{2 \left| \hat{y}_{c,i,k} - y_{c,i,k} \right|}
    {\left| y_{c,i,k} \right| + \left| \hat{y}_{c,i,k} \right|}.
\label{eq:sMAPE}
\end{equation}
The reported weighted sMAPE is computed similar to (\ref{eq:WeightedMRE}) by replacing MRE with sMAPE.

\section{Experimental Results}
\label{sec:V}

The experimental evaluation examines four complementary aspects of the proposed framework, each using the training/validation/test ratio best suited to the quantity under study: 
(i) predictive accuracy under topology-specific training (30\%/35\%/35\%), 
(ii) robustness of the RF-informed inductive bias under intra-class joint training and architectural ablations (30\%/35\%/35\%),  
\new{(iii) comparison with prior art in both predictive accuracy and data efficiency (70\%/15\%/15\% and 30\%/35\%/35\%),}
and (iv) cross-topology knowledge transfer through lightweight fine-tuning within a class (70\%/15\%/15\% and 30\%/35\%/35\%). 
Predictive inference latency relative to conventional circuit simulations is also reported.

\new{The test-heavy 30\%/35\%/35\% split is adopted when the reported quantity is the test-set error statistic, ensuring that the reported results reflect the underlying data distribution rather than small-sample artifacts. Examples include the kernel density estimate (KDE)}
\footnote{A KDE is a non-parametric estimate of a probability density function obtained by centering smooth kernel functions (e.g., Gaussian) on every sample and summing their contributions.} 
\new{plots in Subsection~\ref{subsec:TopologySpecific}, the ablation errors in Subsection~\ref{subsec:ablation}, and the retrained comparison in Subsection~\ref{subsec:SoTA}. The 70\%/15\%/15\% split is adopted for direct comparison with the published FALCON \cite{g19} configuration. The exact pool sizes are provided in the captions of ~\ref{fig:PA}--\ref{fig:Mixer}.}

Collectively, these experiments evaluate the modeling fidelity, representation stability, generalization capability, and deployment efficiency of the proposed surrogate.


\subsection{Topology-Specific Results}
\label{subsec:TopologySpecific}

The proposed framework is trained on one topology and tested on different sizing values of the same topology. The empirical distributions of ground-truth FoMs obtained from Spectre simulations and those predicted by the surrogate are presented as KDE plots in Figures~\ref{fig:PA}--\ref{fig:Mixer}. 
Across all circuit topologies, the predicted densities closely follow the ground-truth distributions, demonstrating the model’s ability to capture heavy-tailed, skewed, and multimodal FoM characteristics typical in nonlinear RF circuits. Quantitatively, the proposed framework achieves a weighted MRE of 1.72\% under topology-specific training, confirming that terminal-level graph representations enable faithful modeling of both device-level (e.g., NF) and block-level (e.g., phase noise, tuning range) circuit behaviors. The detailed performance is presented in Tables~\ref{tab:PA}--\ref{tab:Mixer}.

\begin{table}[b!]
\setlength{\tabcolsep}{2pt}
\vspace{-10pt}
\caption{Average Performance on Unseen Power Amplifier Sizes}
\label{tab:PA}
\centering
\footnotesize
    \begin{tabular}{L{1.3cm}R{1cm}R{0.9cm}R{0.9cm}R{0.9cm}R{0.9cm}R{0.9cm}R{0.9cm}}
    \hline
    {\textbf{Benchmark}} & \textbf{P\textsubscript{DC}} & \textbf{S\textsubscript{11}} & \textbf{S\textsubscript{22}} & \textbf{P\textsubscript{Gain}} & \textbf{DE} & \textbf{PAE} & \textbf{P\textsubscript{SAT}} \\
    \hline
    \textbf{Unit} & \textbf{mW} & \textbf{dB} & \textbf{dB} & \textbf{dB} & \textbf{\%} & \textbf{\%} & \textbf{dBm} \\
    \hline
    MRE & 0.11\% & 1.35\% & 1.07\% & 1.10\% & 1.04\% & 2.93\% & 2.97\% \\ \hline
    sMAPE & 0.11\% & 1.36\% & 1.07\% & 1.04\% & 1.04\% & 2.49\% & 2.22\% \\ \hline
    RMSE & 0.00 & 0.27 & 0.21 & 0.07 & 0.11 & 0.08 & 0.06 \\ \hline
    \end{tabular}
\includegraphics[width=0.73\linewidth]{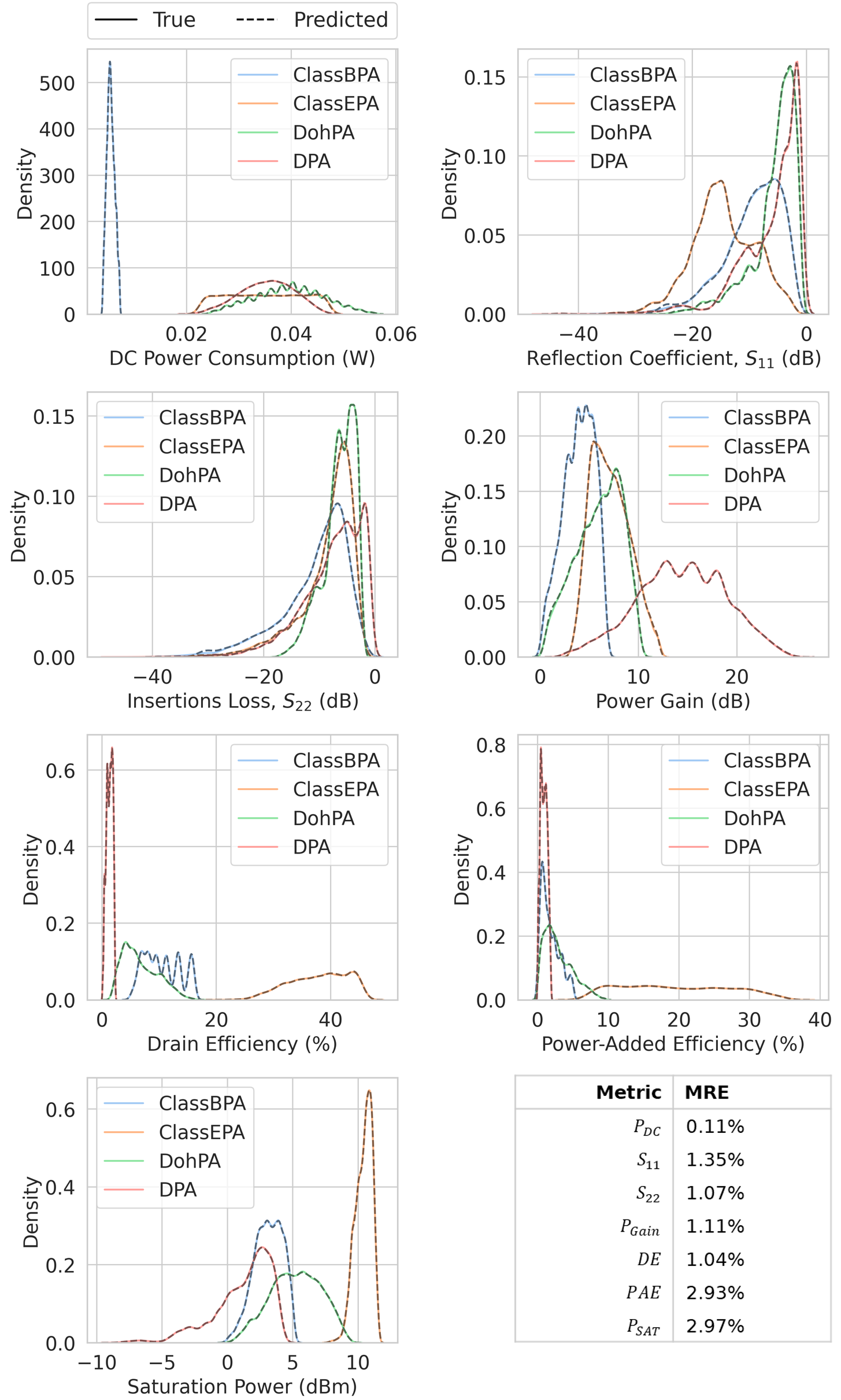}
\captionof{figure}{KDE plots of unseen power amplifier sizes. Average training size: 21{,}087 samples per topology.}
\label{fig:PA}
\end{table}
\begin{table*}[tp]
\setlength{\tabcolsep}{2pt}
\centering
\footnotesize
\begin{minipage}[t]{0.47\textwidth}
\centering
\caption{Average Performance on Unseen Voltage-Controlled Oscillator Sizes}
\label{tab:VCO}
\resizebox{\linewidth}{!}{
\begin{tabular}{L{1.3cm}R{1.3cm}R{1.3cm}R{1.3cm}R{1.3cm}R{1.3cm}}
    \hline
        \textbf{Benchmark} & \textbf{P\textsubscript{DC}} & \textbf{f\textsubscript{osc}} & \textbf{TR} & \textbf{P\textsubscript{out}} & \textbf{PN} \\
        \hline
        \textbf{Unit} & \textbf{mW} & \textbf{GHz} & \textbf{GHz} & \textbf{dBm} & \textbf{dBc/Hz} \\
        \hline
        MRE & 0.11\% & 0.45\% & 4.98\% & 4.10\% & 1.09\% \\ \hline
        sMAPE & 0.11\% & 0.45\% & 3.92\% & 1.81\% & 1.07\% \\ \hline
        RMSE & 0.00 & 0.52 & 0.28 & 0.83 & 2.35 \\ \hline
    \end{tabular}}
    \includegraphics[width=1\linewidth]{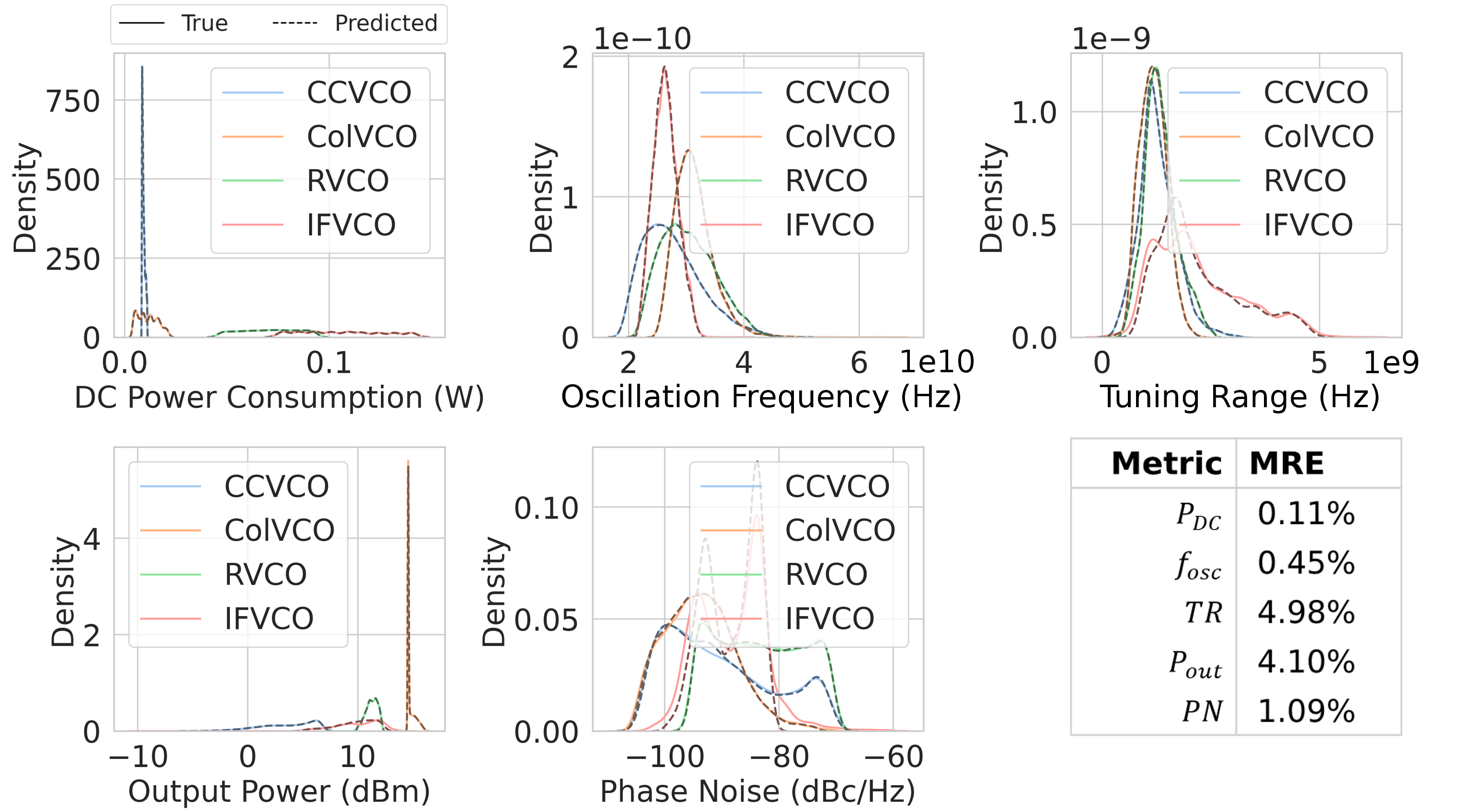}
    \captionof{figure}{KDE Plots of unseen voltage-controlled oscillator sizes. Average training size: 17{,}482 samples per topology.}
    \label{fig:VCO}
\end{minipage}
\hfill
\begin{minipage}[t]{0.47\textwidth}
\centering
\caption{Average Performance on Unseen Voltage Amplifier Sizes}
\label{tab:VA}
\resizebox{\linewidth}{!}{
\begin{tabular}{L{2cm}R{2cm}R{2cm}R{2cm}}
    \hline
        \textbf{Benchmark} & \textbf{BW} & \textbf{P\textsubscript{DC}} & \textbf{V\textsubscript{Gain}} \\
        \hline
        \textbf{Unit} & \textbf{GHz} & \textbf{mW} & \textbf{dB} \\
        \hline
        MRE & 1.59\% & 2.03\% & 4.56\% \\ \hline
        sMAPE & 1.60\% & 2.14\% & 2.99\% \\ \hline
        RMSE & 0.01 & 0.00 & 0.18 \\ \hline
    \end{tabular}}
    \includegraphics[width=0.67\linewidth]{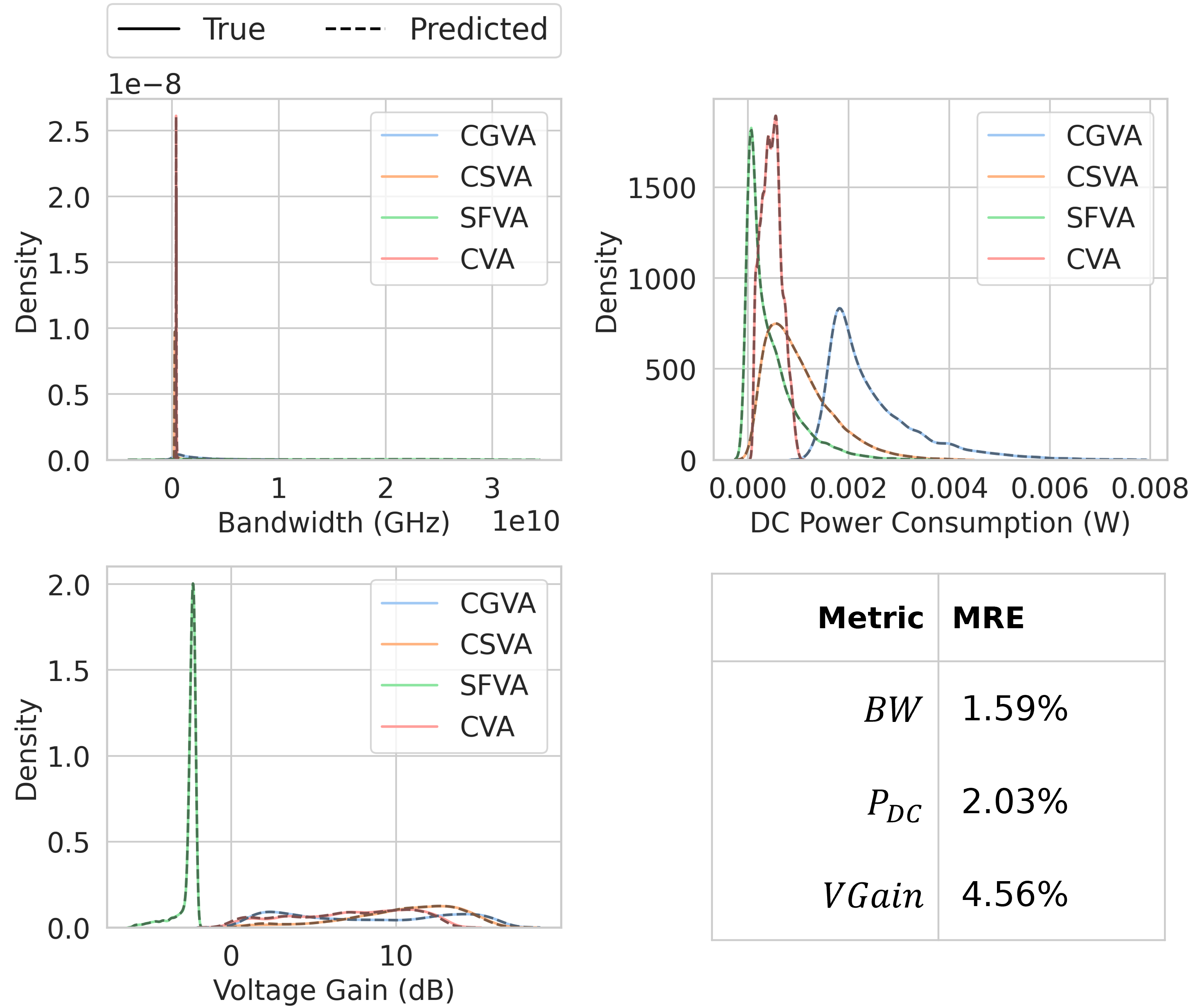}
    \captionof{figure}{KDE plots of unseen voltage amplifier sizes. Average training size: 7{,}862 samples per topology.}
    \label{fig:VA}
\end{minipage}
\par\vspace{1em}
\begin{minipage}[t]{0.47\textwidth}
\centering
\caption{\hspace{-2pt}Average Performance on Unseen Low Noise Amplifier Sizes}
\label{tab:LNA}
\resizebox{\linewidth}{!}{
\begin{tabular}
    {L{1.3cm}R{1.3cm}R{1.3cm}R{1.3cm}R{1.3cm}R{1.3cm}}
    \hline
        \textbf{Benchmark} & \textbf{BW} & \textbf{P\textsubscript{DC}} & \textbf{S\textsubscript{11}} & \textbf{NF} & \textbf{P\textsubscript{Gain}} \\
        \hline
        \textbf{Unit} & \textbf{GHz} & \textbf{mW} & \textbf{dB} & \textbf{dB} & \textbf{dB} \\
        \hline
        MRE & 0.74\% & 0.74\% & 2.66\% & 0.38\% & 3.22\% \\ \hline
        sMAPE & 0.74\% & 0.71\% & 2.64\% & 0.38\% & 2.41\% \\ \hline
        RMSE & 0.10 & 0.00 & 0.94 & 0.04 & 0.09 \\ \hline
        \end{tabular}}
    \includegraphics[width=1\linewidth]{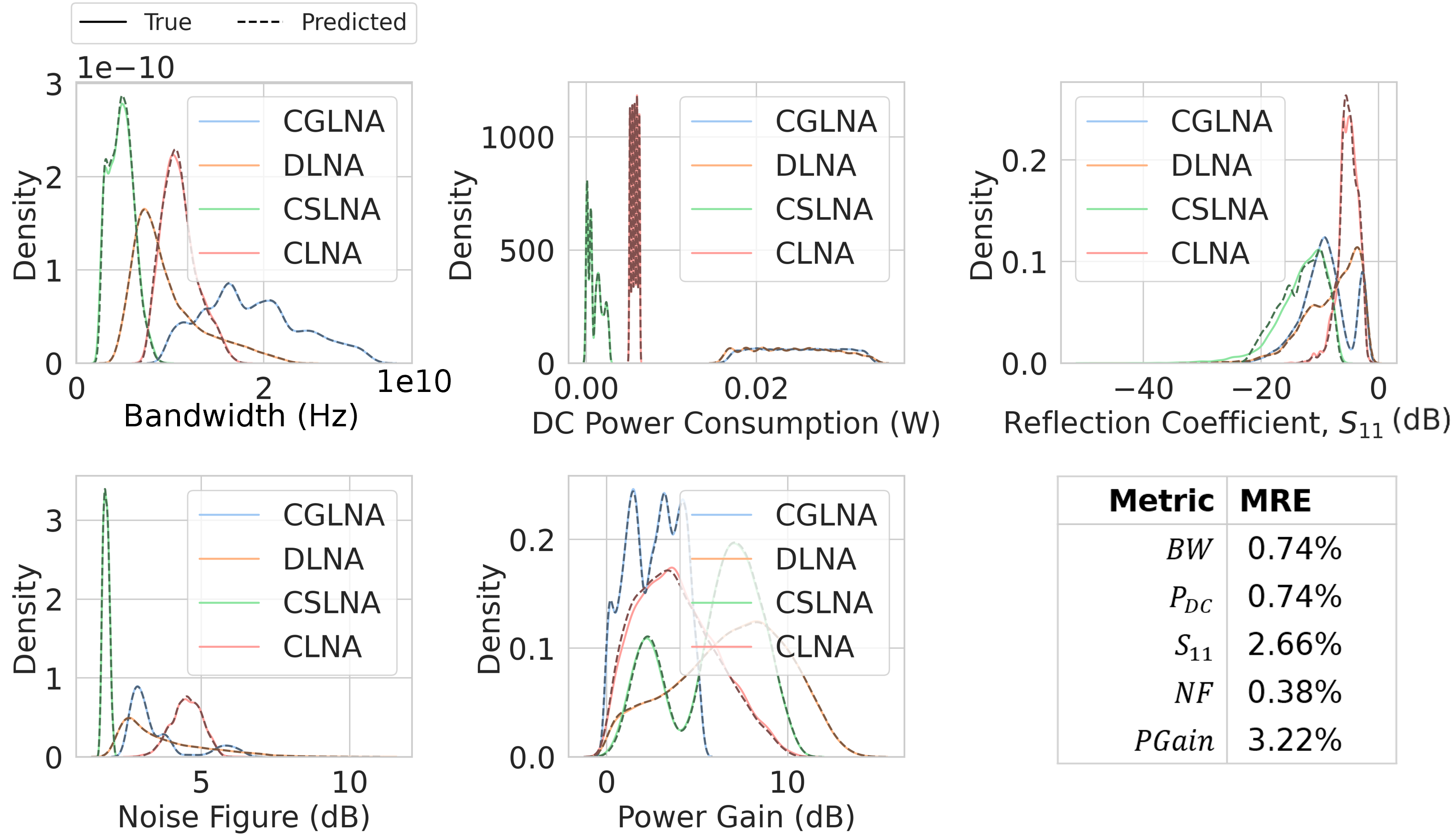}
    \captionof{figure}{KDE plots of unseen low noise amplifier sizes. Average training size: 14{,}828 samples per topology.}
    \label{fig:LNA}
\end{minipage}
\hfill
\begin{minipage}[t]{0.47\textwidth}
\centering
\caption{Average Performance on Unseen Mixer Sizes}
\label{tab:Mixer}
\resizebox{\linewidth}{!}{
\begin{tabular}
    {L{2.2cm}R{1.45cm}R{1.45cm}R{1.45cm}R{1.45cm}}
    \hline
        \textbf{Benchmark} & \textbf{P\textsubscript{DC}} & \textbf{C\textsubscript{Gain}} & \textbf{NF} & \textbf{V\textsubscript{swg}} \\
        \hline
        \textbf{Unit} & \textbf{mW} & \textbf{dB} & \textbf{dB} & \textbf{mV} \\
        \hline
        MRE & 0.19\% & 3.90\% & 0.08\% & 0.59\% \\ \hline
        sMAPE & 0.19\% & 1.37\% & 0.08\% & 0.59\% \\ \hline
        RMSE & 0.00 & 0.93 & 0.01 & 0.00 \\ \hline
    \end{tabular}}
    \includegraphics[width=0.68\linewidth]{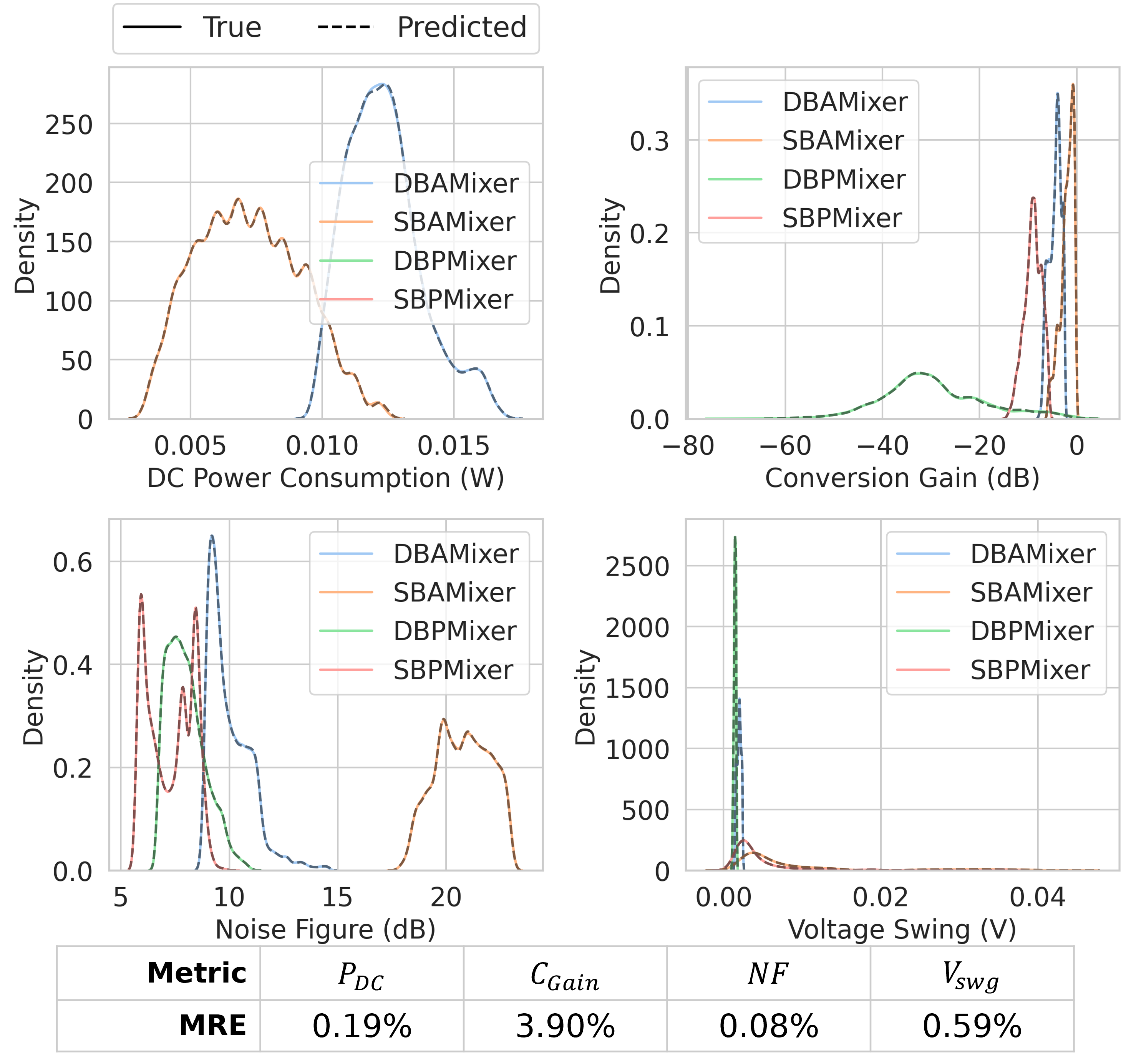}
    \captionof{figure}{KDE plots of unseen mixer sizes. Average training size: 13{,}481 samples per topology.}
    \label{fig:Mixer}
\end{minipage}
\end{table*}

\subsection{Intra-Class Ablation Study}
\label{subsec:ablation}

To evaluate the effectiveness of the proposed RF-informed inductive bias,
the accuracy of the proposed framework is compared with modified variants employing alternative circuit embedding and normalization schemes (Table~\ref{tab:Ablations}). The evaluated variants include: (i) the proposed functionality-aware terminal-level representation with GraphNorm, (ii) functionality-agnostic random feature indexing that ignores the functional roles of individual circuit components, (iii) component-level abstraction, (iv) LayerNorm, and (v) BatchNorm. In component-level abstraction, the circuit components (e.g., transistors, passive devices, power sources) are embedded as the individual graph nodes. Edges are constructed between component nodes whenever the corresponding components share an electrical connection through a common net. Similar to the terminal-level abstraction, component-level node features preserve device parameters such as transistor width, length, and passive element values. However, they omit explicit terminal identity (e.g., source or drain in transistors, negative or positive pole of a voltage source).

All models in this section are trained on all circuit topologies within a given RFIC class and evaluated on the same class using unseen sizing values, with the same training dataset sizes reported in Figures~\ref{fig:PA}--\ref{fig:Mixer}. The performance of the proposed framework is compared against a retrained state-of-the-art model, FALCON \cite{g19}. Since FALCON does not report per-class performance, the original model is retrained separately for each RFIC class using the same dataset sizes to ensure a fair comparison (\textit{reduced per-class} derivation, also defined in Subsection.~\ref{subsec:SoTA}). 

\textbf{Functionality-awareness:}
Across all circuit classes, removing functionality-aware feature indexing leads to substantial degradation in prediction accuracy, \new{dropping the average weighted MRE from 3.45\% to 47.18\% on the same dataset}. This confirms that consistent and 
\new{RF-aware}
encoding of device roles is critical for learning circuit representations. 

\textbf{Graph abstraction:}
Changing the graph abstraction from terminal- to component-level degrades accuracy, with the extent of degradation depending on the underlying physical sensitivity of each FoM. Metrics governed by local current flow, impedance transformation, and device-level matching (e.g., P\textsubscript{Gain} and BW) exhibit larger degradation, as these behaviors rely on fine-grained terminal interactions (e.g., bias asymmetry and parasitic loading) that are not explicitly preserved in component-level abstraction.

Compared to net-level abstraction such as FALCON \cite{g19}, both terminal-level and component-level representations achieve higher accuracy for FoMs dominated by device-level signal transformations (e.g., S\textsubscript{11}, PAE, and NF). However, both underperform net-level abstraction for C\textsubscript{Gain}, a voltage-transfer metric defined between electrical nodes. In terminal-level representations, voltage nets are modeled implicitly as collections of terminals rather than explicit nodes, requiring the GNN to infer voltage relationships indirectly. The results indicate that prediction accuracy is primarily governed by how well the graph abstraction aligns with the dominant physical quantity of the target FoM.

\textbf{Normalization:}
Normalization ablations demonstrate comparable performance between GraphNorm, LayerNorm, and BatchNorm. LayerNorm achieves slightly lower error in smaller or more structurally uniform circuits such as VAs, as it preserves relative differences among devices by normalizing features independently at each node. GraphNorm, in contrast, normalizes features at the graph level, improving robustness when training across circuits with varying graph sizes and connectivity patterns, such as LNAs and PAs. Overall, the differences are modest, indicating that the primary accuracy improvement arises from the functionality-aware feature engineering.

\begin{table}[!t]
\centering
\caption{Average Test MRE \new{in the Reduced Per-Class Setting with Dataset Sizes Similar to Subsection\ref{subsec:TopologySpecific}}.}
\vspace{-5pt}
\label{tab:Ablations}
\footnotesize
\captionsetup[subfloat]{labelformat=empty}
\vspace{-10pt}
\subfloat[]{
\begin{minipage}{\linewidth}
\centering
    \scriptsize
    \adjustbox{width=\linewidth}{
    \setlength{\tabcolsep}{3pt}
    \begin{tabular}{llrrrrrrrr}
    \multicolumn{9}{c}{Average Test MRE (\%) Across the PA Class} \\
    \toprule
    \textbf{Set} & \textbf{P\textsubscript{DC}} & \textbf{S\textsubscript{11}} & \textbf{S\textsubscript{22}} & \textbf{P\textsubscript{Gain}} & \textbf{DE} & \textbf{PAE} & \textbf{P\textsubscript{SAT}} & \textbf{Avg.} \\
    \hline
    \textbf{This~work} & \textbf{0.1} & 1.7 & 1.5 & 2.3 & \textbf{1.5} & \textbf{5.7} & \textbf{2.7} & \textbf{2.2} \\ \hline
    Random features & 9.5 & 64.7 & 55.9 & 40.9 & 37.0 & 99.8 & 93.8 & 57.4 \\ \hline
    Component-based & 0.1 & 2.0 & 2.0 & 2.2 & 2.5 & 5.8 & 3.5 & 2.6 \\ \hline
    LayerNorm & 0.1 & \textbf{1.6} & \textbf{1.4} & \textbf{2.2} & 1.8 & 6.5 & 3.4 & 2.4 \\ \hline
    BatchNorm & 0.1 & 2.5 & 1.9 & 2.4 & 3.0 & 8.9 & 4.5 & 3.3 \\ \hline
    FALCON & 1.7 & 22.8 & 22.6 & 14.0 & 28.2 & 64.7 & 31.9 & 26.6 \\
\end{tabular}}
\end{minipage}}
\vspace{-20pt}
\subfloat[]{
\begin{minipage}{\linewidth}
\centering
    \adjustbox{width=\linewidth}{%
    \setlength{\tabcolsep}{3pt}
    \begin{tabular}{L{2.1cm}L{0.6cm}R{1.1cm}R{1.1cm}R{1.1cm}R{1cm}}
    \toprule
    \multicolumn{6}{c}{Average Test MRE (\%) Across the Mixer Class} \\
    \midrule
    \textbf{Set} & \textbf{P\textsubscript{DC}} & \textbf{C\textsubscript{Gain}} & \textbf{NF} & \textbf{V\textsubscript{swg}} & \textbf{Avg.} \\
    \hline
    \textbf{This~work} & \textbf{0.4} & 14.9 & \textbf{0.8} & \textbf{2.0} & 4.5 \\ \hline
    Random features & 2.9 & 64.6 & 6.6 & 49.7 & 31.0 \\ \hline
    Component-based & 0.7 & 18.6 & 1.2 & 2.1 & 5.6 \\ \hline
    LayerNorm & 0.4 & 15.2 & 0.8 & 2.1 & 4.6 \\ \hline
    BatchNorm & 0.4 & 17.2 & 0.9 & 2.1 & 5.1 \\ \hline
    FALCON & 0.6 & \textbf{11.1} & 1.8 & 4.0 & \textbf{4.4} \\
\end{tabular}}
\end{minipage}}
\vspace{-20pt}

\subfloat[]{
\begin{minipage}{\linewidth}
    \adjustbox{width=\linewidth}{%
    \setlength{\tabcolsep}{3pt}
\begin{tabular}{L{2.2cm}L{0.7cm}R{0.7cm}R{0.9cm}R{0.8cm}R{0.8cm}R{0.8cm}}
\toprule
\multicolumn{7}{c}{Average Test MRE (\%) Across the VCO Class} \\
\midrule
    \textbf{Set} & \textbf{P\textsubscript{DC}} & \textbf{f\textsubscript{osc}} & \textbf{TR} & \textbf{P\textsubscript{out}} & \textbf{PN} & \textbf{Avg.} \\
    \hline
    \textbf{This~work} & \textbf{0.2} & \textbf{0.7} & \textbf{5.4} & \textbf{4.4} & \textbf{1.2} & \textbf{2.4} \\ \hline
    Random features & 5.6 & 8.5 & 23.5 & 83.8 & 5.5 & 25.4 \\ \hline
    Component-based & 0.3 & 1.1 & 6.2 & 5.1 & 1.4 & 2.8 \\ \hline
    LayerNorm & 0.2 & 0.7 & 5.5 & 4.5 & 1.2 & 2.4 \\ \hline
    BatchNorm & 0.3 & 1.0 & 5.9 & 4.7 & 1.3 & 2.6 \\ \hline
    FALCON & 5.2 & 1.1 & 6.9 & 5.1 & 1.6 & 4.0 \\
\end{tabular}}
\end{minipage}}

\vspace{-20pt}
\subfloat[]{
\begin{minipage}{\linewidth}
    \adjustbox{width=\linewidth}{%
    \setlength{\tabcolsep}{3pt}
\begin{tabular}{L{2.1cm}R{0.5cm}R{0.9cm}R{0.8cm}R{0.9cm}R{0.9cm}R{0.7cm}}
\toprule
\multicolumn{7}{c}{Average Test MRE (\%) Across the LNA Class} \\
\midrule
    \textbf{Set} & \textbf{BW} & \textbf{P\textsubscript{DC}} & \textbf{S\textsubscript{11}} & \textbf{NF} & \textbf{P\textsubscript{Gain}} & \textbf{Avg.} \\
    \hline
    \textbf{This~work} & \textbf{2.4} & \textbf{1.2} & \textbf{6.5} & \textbf{1.6} & \textbf{15.1} & \textbf{5.4} \\ \hline
    Random features & 19.6 & 3.7 & 42.2 & 14.5 & 215.7 & 59.2 \\ \hline
    Component-based & 9.9 & 2.3 & 9.5 & 2.1 & 134.3 & 31.6 \\ \hline
    LayerNorm & 2.4 & 1.2 & 6.5 & 1.6 & 16.6 & 5.6 \\ \hline
    BatchNorm & 2.7 & 2.9 & 7.0 & 1.7 & 18.4 & 6.5 \\ \hline
    FALCON & 27.8 & 40.8 & 62.2 & 18.3 & 343.3 & 98.5 \\
\end{tabular}}
\end{minipage}}

\vspace{-20pt}
\subfloat[]{
\begin{minipage}{\linewidth}
    \adjustbox{width=\linewidth}{%
    \setlength{\tabcolsep}{3pt}
\begin{tabular}{L{2.1cm}R{0.7cm}R{1.5cm}R{1.5cm}R{1.5cm}}
\toprule
\multicolumn{5}{c}{Average Test MRE (\%) Across the VA Class} \\
\midrule
    \textbf{Set} & \textbf{BW} & \textbf{P\textsubscript{DC}} & \textbf{V\textsubscript{Gain}} & \textbf{Avg.} \\ \hline
    \textbf{This~work} & 3.9 & 2.2 & 11.5 & 5.8 \\ \hline
    Random features & 35.7 & 29.0 & 88.2 & 51.0 \\ \hline
    Component-based & 5.8 & 4.1 & 19.5 & 9.8 \\ \hline
    LayerNorm & \textbf{3.2} & \textbf{1.9} & \textbf{9.7} & \textbf{5.0} \\ \hline
    BatchNorm & 6.1 & 3.8 & 54.1 & 21.3 \\ \hline
    FALCON & 36.1 & 14.9 & 25.7 & 25.6 \\
\bottomrule
\end{tabular}}
\end{minipage}}
\end{table}

\begin{table*}[t]
    \setlength{\tabcolsep}{2pt}
    \vspace{-10pt}
    \centering
    \caption{Per-Metric sMAPE of This Work and FALCON~\cite{g19} Trained on the Same Dataset of 300k.}
    \label{tab:sMAPE}
    \small
    \renewcommand{\arraystretch}{1.3}
    \resizebox{0.9\textwidth}{!}{%
    \begin{tabular}{c||c||cccccccccccccccc|c}
    \toprule
    \textbf{Work} & \textbf{Dataset} & 
    \multicolumn{16}{c|}{\textbf{sMAPE (\%) per Performance Metric}} &
    Weighted \\
    \cmidrule(lr){3-18}
    
    & & P\textsubscript{DC} & V\textsubscript{Gain} & P\textsubscript{Gain} & C\textsubscript{Gain} & S\textsubscript{11} & S\textsubscript{22} & NF & BW & f\textsubscript{osc} & TR & P\textsubscript{out} & P\textsubscript{SAT} &
    DE & PAE & PN & V\textsubscript{swg} & Avg. (\%) \\ \midrule
    
    This work & Reduced per-class & 0.6 & 8.0 & 4.9 & 6.1 & 3.7 & 1.5 & 1.2 & 2.9 & 0.7 & 4.1 & 2.1 & 2.0 & 1.6 & 4.9 & 1.2 & 2.0 & 2.57 \\ \hline
    
    FALCON~\cite{g19} & Reduced & 10.2 & 5.8 & 8.6 & 6.6 & 11.5 & 4.5 & 5.1 & 11.1 & 0.9 & 7.6 & 5.6 & 7.2 & 11.0 & 17.7 & 1.6 & 3.2 & 8.19 \\ \bottomrule
    \end{tabular}}
\end{table*}

\begin{figure}[!htbp]
  \centering
  \includegraphics[width=\columnwidth]{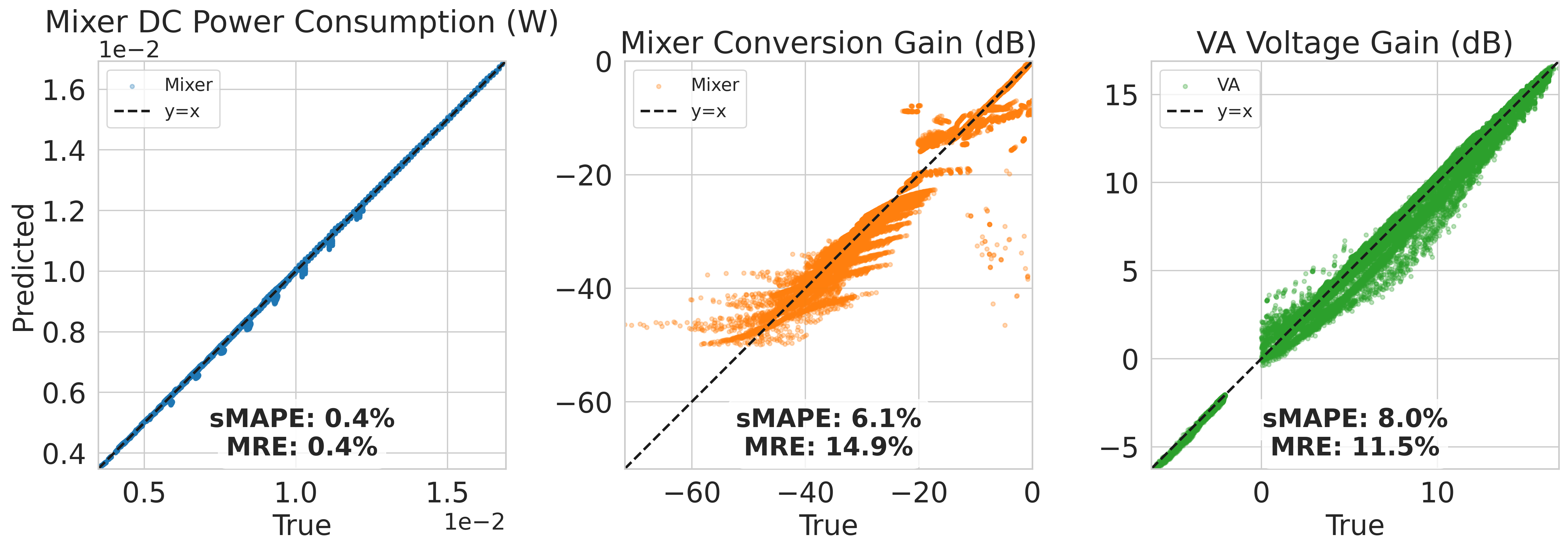}
  \caption{Parity plots, MRE, and sMAPE for a well-performing FoM, mixer P\textsubscript{DC} (blue shade), and two worse-performing FoMs, mixer C\textsubscript{Gain} (orange shade) and VA V\textsubscript{Gain} (green shade).}
  \label{fig:ParityPlots}
\end{figure}

\begin{table}[!htbp]
\setlength{\tabcolsep}{3pt}
\renewcommand{\arraystretch}{0.6}
\vspace{-5pt}
\centering
\caption{Average Test MRE (\%) for Knowledge Transfer for Complete, Per-Class (P.C.), and Reduced Per-Class (R.P.C.) Settings}
\vspace{-10pt}
\label{tab:FineTune}
\footnotesize
\captionsetup[subfloat]{labelformat=empty}
\vspace{-10pt}
\subfloat[]{
\begin{minipage}{\linewidth}
\centering
    \begin{tabularx}{\linewidth}{p{1.1cm}|p{1.1cm}|p{0.7cm}|*{7}{Y}|Z}
    \toprule
        \multicolumn{11}{c}{\scriptsize Knowledge Transfer from ClassBPA, DPA, and DohPA to ClassEPA} \\ \midrule
        \textbf{Set} & \textbf{Dataset} & \textbf{Layer} & \textbf{P\textsubscript{DC}} & \textbf{S\textsubscript{11}} & \textbf{S\textsubscript{22}} & \textbf{P\textsubscript{Gain}} & \textbf{DE} & \textbf{PAE} & \textbf{P\textsubscript{SAT}} & \textbf{Avg.} \\ \midrule
        
        \textbf{This~work} & \textbf{P.C.} & \textbf{1\textsuperscript{st}} & \textbf{0.1} & \textbf{0.6} & \textbf{1.4} & \textbf{0.2} & \textbf{0.4} & \textbf{0.7} & \textbf{0.1} & \textbf{0.5} \\
        
        This~work & P.C. & 2\textsuperscript{nd} & 0.1 & 0.7 & 1.8 & 0.3 & 0.6 & 0.8 & 0.1 & 0.6 \\
        
        This~work & P.C. & 3\textsuperscript{rd} & 0.1 & 1.0 & 2.8 & 0.5 & 1.0 & 1.5 & 0.3 & 1.0 \\
        
        This~work & P.C. & 4\textsuperscript{th} & 0.2 & 1.8 & 7.1 & 1.4 & 3.2 & 3.3 & 0.8 & 2.5 \\
        
        This~work & P.C. & Last & 0.6 & 15.9 & 27.1 & 6.5 & 9.0 & 11.0 & 1.8 & 10.3 \\
        This~work & R.P.C. & 1\textsuperscript{st} & 0.1 & 0.7 & 1.9 & 0.3 & 0.6 & 0.9 & 0.1 & 0.7 \\
        FALCON & R.P.C. & Last & 7.4 & 29.7 & 45.5 & 14.3 & 30.5 & 38.9 & 16.3 & 26.1 \\
        
        FALCON & Reduced & Last & 5.7 & 6.4 & 6.4 & 7.6 & 3.9 & 8.7 & 3.1 & 7.0 \\
        
        FALCON & Complete & Last & 4.4 & 5.6 & 11.7 & 5.3 & 2.4 & 5.4 & 3.9 & 5.5 \\
        
        FALCON & Complete & 1\textsuperscript{st} & 27.9 & 45.1 & 40.6 & 19.7 & 78.8 & 90.6 & 69.4 & 53.2 \\
    \end{tabularx}
\end{minipage}}
\vspace{-20pt}
\subfloat[]{
\begin{minipage}{\linewidth}
\centering
    \begin{tabularx}{\linewidth}{p{1.1cm}|p{1.1cm}|p{0.7cm}|*{4}{Y}|Z}
    \toprule
        \multicolumn{8}{c}{\scriptsize Knowledge~Transfer from SBPMixer, SBAMixer, and DBPMixer to DBAMixer} \\ \midrule
        \textbf{Set} & \textbf{Dataset} & \textbf{Layer} & \textbf{P\textsubscript{DC}} & \textbf{C\textsubscript{Gain}} & \textbf{NF} & \textbf{V\textsubscript{swg}} & \textbf{Avg.} \\ \midrule
        
        \textbf{This~work} & \textbf{P.C.} & \textbf{1\textsuperscript{st}} & \textbf{0.2} & \textbf{0.2} & \textbf{0.1} & \textbf{0.2} & \textbf{0.2} \\ 
        
        This~work & P.C. & 2\textsuperscript{nd} & 0.2 & 0.4 & 0.1 & 0.2 & 0.2 \\ 
        
        This~work & P.C. & 3\textsuperscript{rd} & 0.2 & 0.6 & 0.1 & 0.2 & 0.3 \\ 
        
        This~work & P.C. & 4\textsuperscript{th} & 0.2 & 0.8 & 0.2 & 0.3 & 0.4 \\ 
        
        This~work & P.C. & Last & 1.4 & 6.5 & 1.8 & 3.2 & 3.2 \\
        This~work & R.P.C. & 1\textsuperscript{st} & 0.2 & 0.5 & 0.2 & 0.2 & 0.3 \\
        FALCON & R.P.C. & Last & 4.3 & 32.2 & 5.6 & 32.5 & 18.7 \\
        
        FALCON & Reduced & Last & 28.2 & 115.4 & 20.7 & 247.7 & 103.0 \\
        
        FALCON & Complete & Last & 3.7 & 7.6 & 1.1 & 11.3 & 5.9 \\
        
        FALCON & Complete & 1\textsuperscript{st} & 86.7 & 97.9 & 14.6 & 194.6 & 98.4 \\
    \end{tabularx}
\end{minipage}}
\vspace{-20pt}
\subfloat[]{
\begin{minipage}{\linewidth}
\centering
    \begin{tabularx}{\linewidth}{p{1.1cm}|p{1.1cm}|p{0.7cm}|*{5}{Y}|Z}
    \toprule
        \multicolumn{9}{c}{\scriptsize Knowledge Transfer from IFVCO, CCVCO, and ColVCO to RVCO} \\ \midrule
        \textbf{Set} & \textbf{Dataset} & \textbf{Layer} & \textbf{P\textsubscript{DC}} & \textbf{f\textsubscript{osc}} & \textbf{TR} & \textbf{P\textsubscript{out}} & \textbf{PN} & \textbf{Avg.} \\ \midrule
        
        \textbf{This~work} & \textbf{P.C.} & \textbf{1\textsuperscript{st}} & \textbf{0.1} & \textbf{0.1} & \textbf{0.4} & \textbf{0.1} & \textbf{0.3} & \textbf{0.2} \\
        
        This~work & P.C. & 2\textsuperscript{nd} & 0.1 & 0.2 & 0.4 & 0.1 & 0.3 & 0.2 \\
        
        This~work & P.C. & 3\textsuperscript{rd} & 0.1 & 0.2 & 0.4 & 0.1 & 0.3 & 0.2 \\
        
        This~work & P.C. & 4\textsuperscript{th} & 0.1 & 0.2 & 0.5 & 0.1 & 0.3 & 0.2 \\
        
        This~work & P.C. & Last & 0.9 & 1.6 & 4.8 & 1.0 & 1.7 & 2.0 \\
        This~work & R.P.C. & 1\textsuperscript{st} & 0.1 & 0.2 & 0.5 & 0.0 & 0.3 & 0.2 \\
        FALCON & R.P.C. & Last & 3.1 & 2.5 & 4.5 & 2.2 & 1.4 & 2.7 \\
        
        FALCON & Reduced & Last & 1.1 & 0.6 & 2.1 & 1.2 & 0.8 & 1.2 \\
        
        FALCON & Complete & Last & 1.0 & 0.7 & 1.8 & 1.0 & 0.6 & 1.0 \\
        
        FALCON & Complete & 1\textsuperscript{st} & 57.0 & 7.6 & 24.8 & 9.5 & 5.8 & 20.9 \\
    \end{tabularx}
\end{minipage}}
\vspace{-20pt}
\subfloat[]{
\begin{minipage}{\linewidth}
\centering
    \begin{tabularx}{\linewidth}{p{1.1cm}|p{1.1cm}|p{0.7cm}|*{5}{Y}|Z}
    \toprule
        \multicolumn{9}{c}{\scriptsize Knowledge Transfer from CGLNA, CLNA, and DLNA to CSLNA} \\ \midrule
        \textbf{Set} & \textbf{Dataset} & \textbf{Layer} & \textbf{BW} & \textbf{P\textsubscript{DC}} & \textbf{S\textsubscript{11}} & \textbf{NF} & \textbf{P\textsubscript{Gain}} & \textbf{Avg.} \\ \midrule
        
        \textbf{This~work} & \textbf{P.C.} & \textbf{1\textsuperscript{st}} & \textbf{1.0} & \textbf{0.9} & \textbf{7.2} & \textbf{0.2} & \textbf{1.6} & \textbf{2.2} \\
        
        This~work & P.C. & 2\textsuperscript{nd} & 1.0 & 2.0 & 7.3 & 0.3 & 1.7 & 2.5 \\
        
        This~work & P.C. & 3\textsuperscript{rd} & 1.0 & 4.7 & 7.4 & 0.3 & 2.0 & 3.1 \\
        
        This~work & P.C. & 4\textsuperscript{th} & 1.8 & 9.3 & 10.6 & 0.4 & 7.4 & 5.9 \\
        
        This~work & P.C. & Last & 18.7 & 191.9 & 19.9 & 2.3 & 76.3 & 61.8 \\
        This~work & R.P.C. & 1\textsuperscript{st} & 1.0 & 2.1 & 7.2 & 0.2 & 1.6 & 2.4 \\
        FALCON & R.P.C. & Last & 198.7 & 4099.6 & 24.7 & 43.1 & 136.9 & 903.28 \\
        
        FALCON & Reduced & Last & 3.2 & 493.8 & 15.1 & 21.4 & 22.2 & 113.2 \\
        
        FALCON & Complete & Last & 7.6 & 360.5 & 10.4 & 11.5 & 6.7 & 79.4 \\
        
        FALCON & Complete & 1\textsuperscript{st} & 125.5 & 7660.7 & 32.0 & 362.8 & 40.0 & 1644.7 \\
    \end{tabularx}
\end{minipage}}
\vspace{-20pt}
\subfloat[]{
\begin{minipage}{\linewidth}
\centering
    \begin{tabularx}{\linewidth}{p{1.1cm}|p{1.1cm}|p{0.7cm}|*{3}{Y}|Z}
    \toprule
        \multicolumn{7}{c}{\scriptsize Knowledge Transfer from CSVA, CGVA, and SFVA to CVA} \\ \midrule
        \textbf{Set} & \textbf{Dataset} & \textbf{Layer} & \textbf{BW} & \textbf{P\textsubscript{DC}} & \textbf{V\textsubscript{Gain}} & \textbf{Avg.} \\ \midrule
        
        This~work & P.C. & 1\textsuperscript{st} & \textbf{0.1} & \textbf{0.5} & 13.5 & 4.7 \\
        
        This~work & P.C. & 2\textsuperscript{nd} & 0.1 & 0.6 & 11.4 & 4.0 \\
        
        This~work & P.C. & 3\textsuperscript{rd} & 0.2 & 0.5 & 12.9 & 4.5 \\
        
        \textbf{This~work} & \textbf{P.C.} & \textbf{4\textsuperscript{th}} & 0.2 & 0.5 & \textbf{9.49} & \textbf{3.4} \\
        
        This~work & P.C. & Last & 2.5 & 4.9 & 114.0 & 40.4 \\
        This~work & R.P.C. & 1\textsuperscript{st} & 0.2 & 0.6 & 16.2 & 5.7 \\
        FALCON & R.P.C. & Last & 252.8 & 411.2 & 289.5 & 317.9 \\
        
        FALCON & Reduced & Last & 2013.2 & 2001.3 & 143.0 & 1386.0 \\ 
        
        FALCON & Complete & Last & 66.7 & 199.8 & 22.8 & 96.7 \\ 
        
        FALCON & Complete & 1\textsuperscript{st} & 2109.6 & 3383.1 & 119.0 & 1871.5 \\
    \end{tabularx}
\end{minipage}}
\vspace{-20pt}
\subfloat[]{
\begin{minipage}{\linewidth}
\centering
    \begin{tabularx}{\linewidth}{p{1.1cm}|p{1.1cm}|p{0.7cm}|>{\centering\arraybackslash}X}
    \toprule
        \multicolumn{4}{c}{\scriptsize Average Weighted MRE} \\ \midrule
        \textbf{Set} & \textbf{Dataset} & \textbf{Layer} & \textbf{Avg.} \\ \midrule
        
        This~work & P.C. & 1\textsuperscript{st} & \textbf{1.1} \\
        %
        %
        %
        %
        %
        
        FALCON & Complete & Last & 28.8 \\ \bottomrule 
        %
    \end{tabularx}
\end{minipage}}
\end{table}

\textbf{Parity Plots:}
Calibration, bias, and error distribution are illustrated through representative high- and lower-accuracy parity plots in Figure~\ref{fig:ParityPlots}. For well-predicted metrics (e.g., mixer P\textsubscript{DC}), predictions closely follow the $y=x$ line across the full dynamic range, confirming absence of systematic bias. In contrast, mixer C\textsubscript{Gain} and VA V\textsubscript{Gain} exhibit increased scatter (notably in DBAMixer) and localized deviations (notably in SFVA), consistent with the outlier distributions observed in the KDE plots in Tables~\ref{tab:Mixer} and \ref{tab:VA}. Despite the increased variance, predictions remain centered around the $y=x$ line, indicating that the model captures the overall FoM trend but exhibits reduced precision in gain-based ones.

The relatively lower accuracy for gain-related metrics partly attributes to the definition of MRE, which penalizes over-prediction more severely when true values are small (see (\ref{eq:MRE})). Thus, although C\textsubscript{Gain} predictions are generally well aligned with the $y=x$ line, a small number of over-predicted outliers disproportionately increases the MRE. Nevertheless, MRE is used for fair comparison with prior work. To provide additional perspective, a symmetric metric (sMAPE) is also reported in Table~\ref{tab:sMAPE}. Compared to FALCON, the proposed framework achieves lower sMAPE for C\textsubscript{Gain} (6.1\% vs 6.6\%) and slightly higher sMAPE for V\textsubscript{Gain} (8.0\% vs 5.8\%). The results demonstrate improved overall symmetric performance (i.e., no directional bias).

\textbf{Conclusion:}
Overall, the proposed configuration achieves the lowest or near-lowest MRE across all classes and FoMs.
The limited sensitivity to node abstraction and normalization selection indicates that performance improvements primarily stem from the proposed functionality-aware representation rather than fragile architectural tuning.

\subsection{Comparison with State-of-the-Art}
\label{subsec:SoTA}

The performance of the proposed framework is compared to the most comparable work in state-of-the-art with \new{open-source code and data}, FALCON \cite{g19}. 
FALCON is retrained with 30\%/35\%/35\% ratios once on individual RFIC classes (\textit{reduced per-class}, total training size of 60k$\times$5) and once on all circuits jointly (\textit{reduced}, total training size of 300k$\times$1). The original FALCON model is reported as trained on all circuits, except for the held-out RVCO, with ratios of 70\%/15\%/15\% (nineteen circuits, \textit{complete w/o RVCO}, total training size of \new{665k$\times$1}).
\new{This framework is trained with 70\%/15\%/15\% ratios with RVCO (\textit{per-class}, total training size of 140k$\times$5), and without RVCO (nineteen circuits, \textit{per-class w/o RVCO}), total training size of 140k$\times$5$-$35k)}. This work is also trained with 30\%/35\%/35\% ratios (\textit{reduced per-class}, total training size of 60k$\times$5). Details are listed in Table~\ref{tab:accuracy_comparison}.

As mentioned in Subsection~\ref{subsec:ablation}, solutions that map voltage nets to graph nodes such as \cite{g19}, outperform in metrics driven primarily by voltage propagation in loops (e.g., C\textsubscript{Gain} and V\textsubscript{Gain}). In contrast, this work maps device terminals to graph nodes and therefore better captures terminal-level current flow, device operating regions, 
and parasitic behavior,
mostly beneficial for P\textsubscript{DC}, S\textsubscript{11}, P\textsubscript{Gain}, and PAE. As summarized in Table~\ref{tab:accuracy_comparison}, a majority of RF FoMs are influenced by terminal-level device interactions, leading to improved overall electrical accuracy and a lower weighted MRE for the proposed framework.

\new{Unified frameworks such as \cite{g19} offer deployment simplicity, but surprisingly, forcing a unified model to be per-class, increases the MRE from 13.72\% to 31.74\% (a degradation of $\sim$2.3$\times$) while trained with the same dataset size. This work over-performs FALCON in its intended unified setting as well, achieving a weighted MRE of 2.71\% on the same dataset of FALCON's 9.09\% reported weighted MRE, an improvement of 3.3$\times$. Including the relatively simple circuit of RVCO leads to the weighted MRE of 2.60\%. With $\sim$2.3$\times$ less training data, the weighted MRE only increases to 3.45\%. The proposed per-class, RF-aware model enables parallel training of multiple circuit classes across GPUs and removes the overhead of training a costly encoder (explained in detail in Subsection\ref{subsec:TL}). In its \textit{per-class} mode, this work reduces training time 
of FALCON \textit{complete} from 100 hours to 5 hours, a speed-up of 20$\times$. Training time is defined here as training and validation time of all 20 circuits and includes per-batch data fetching and CPU-GPU data transfer time.}

\begin{figure}[t]
  \vspace{-3pt}
  \centering
  \includegraphics[width=\columnwidth]{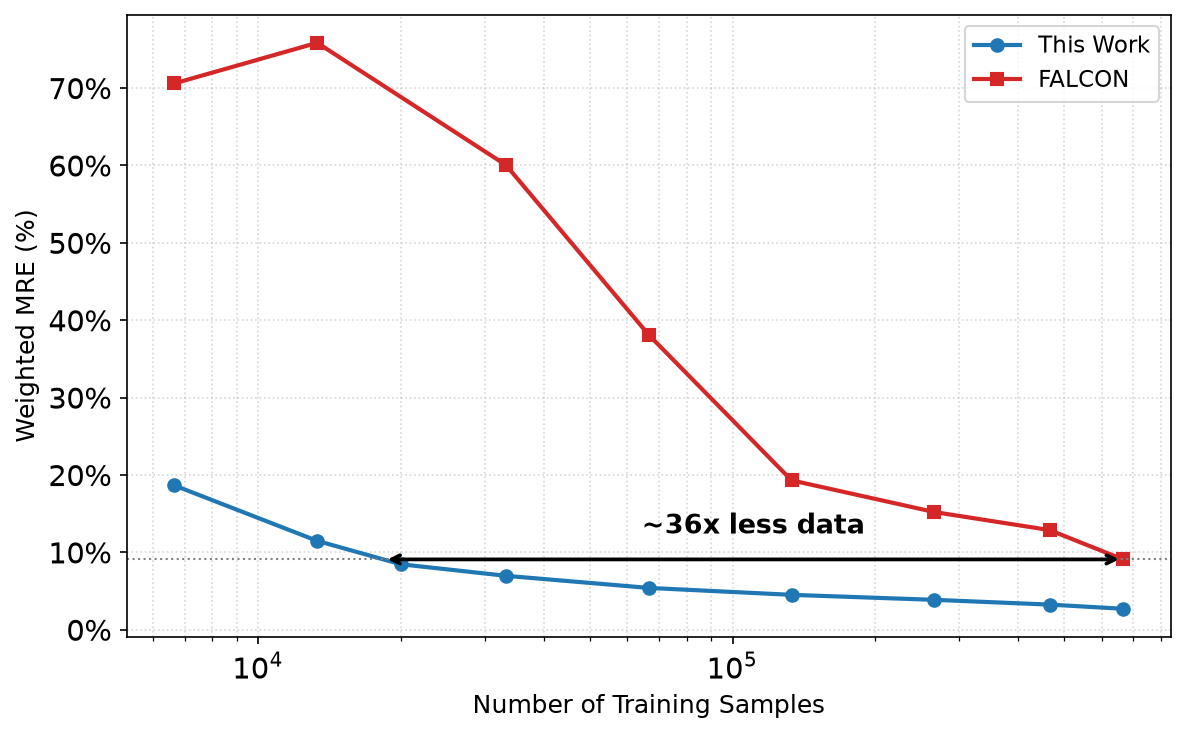}
  \vspace{-15pt}
  \caption{\new{Training/validation/test pools are fixed, and subsets of training pool are selected for training. Compared to FALCON \cite{g19}, this work needs $\sim$36$\times$ less training data to achieve the weighted MRE of $\sim$9.09\% on the test set.}}
  \label{fig:DataEfficiency}
\end{figure}

\new{To showcase data efficiency, training/validation/test pools are fixed between this work and FALCON \cite{g19} to cover 70\%/15\%/15\% of the whole dataset excluding RVCO, as reported in FALCON. In each iteration, the frameworks are trained on randomly chosen subsets of the training pool (i.e., 1\%, 2\%, 5\%, 10\%, 20\%, 40\%, 70\%, 100\%). The MRE results  are shown in Figure~\ref{fig:DataEfficiency}. 
Compared to FALCON's MRE of $\sim$9.09\% using the full (100\%) training pool, the proposed framework reaches  $\sim$18k training samples, corresponding to a $\sim$36$\times$ improvement in data efficiency.}
\new{These results indicate that functionality-aware feature indexing and class-specialized training improve the model’s ability to learn circuit behavior in general while reducing data and computational requirements.}

\begin{table*}[t]
    \setlength{\tabcolsep}{2pt}
    \caption{Per-Metric MRE and Training Efficiency of the Proposed Framework and FALCON~\cite{g19} under Various Settings.}
    \label{tab:accuracy_comparison}
    \small
    \renewcommand{\arraystretch}{1.4}
    \resizebox{\textwidth}{!}{
    \begin{tabular}{c||lccr||cccccccccccccccc|c}
    \toprule
    
    \textbf{Work} & \multicolumn{4}{c||}{Training} & \multicolumn{16}{c}{MRE per Performance Metric (\%)}\vspace{2pt} & \multicolumn{1}{|c}{Weighted} \\
    \cline{2-21}

    & ~Dataset & Models & ~Samples~ & Hours~ & P\textsubscript{DC} & V\textsubscript{Gain} & P\textsubscript{Gain} & C\textsubscript{Gain} & S\textsubscript{11} & S\textsubscript{22} & NF & BW & f\textsubscript{osc} & TR & P\textsubscript{out} & P\textsubscript{SAT} & DE & PAE & PN & V\textsubscript{swg} & Avg. (\%) \\ \midrule

    This~work & ~Per-class & 5 & 140k$\times$5 & 5~
    & 0.5 & 6.1 & 5.0 & 9.6 & 3.2 & 1.1 & 1.0 & 2.5 & 0.7 & 5.2 & 3.8 & 1.7 & 1.1 & 4.8 & 1.2 & 1.9 & 2.60 \\ \hline

    This~work & \makecell[tl]{~Per-class\\~w/o RVCO} & 5 & 140k$\times$5$-$35k & 5~
    & 0.5 & 6.1 & 5.0 & 9.6 & 3.2 & 1.1 & 1.0 & 2.5 & 0.8 & 6.3 & 4.6 & 1.7 & 1.1 & 4.8 & 1.4 & 1.9 & 2.71 \\ \hline
    
    This~work & \makecell[tl]{~Reduced\\~per-class} & 5 & 60k$\times$5 & 4~
    & 0.6 & 11.5 & 7.6 & 14.8 & 3.7 & 1.5 & 1.2 & 2.9 & 0.7 & 5.4 & 4.4 & 2.7 & 1.5 & 5.7 & 1.2 & 2.0 & 3.45 \\ \hline
    
    FALCON & \makecell[tl]{~Complete\\~w/o RVCO} & 1 & \new{665k$\times$1} & $\sim$100~ 
    & 11.2 & 2.6 & 19.0 & 6.1 & 11.4 & 1.9 & 4.5 & 6.5 & 0.6 & 6.5 & 4.6 & 4.4 & 4.6 & 11.0 & 1.3 & 1.4 & 9.09 \\ \hline

    FALCON & ~Reduced & 1 & 300k$\times$1 & $\sim$100~
    & 25.6 & 9.8 & 25.2 & 10.2 & 14.3 & 4.4 & 5.4 & 10.6 & 0.9 & 7.6 & 5.6 & 11.2 & 10.6 & 27.3 & 1.6 & 3.3 & 13.72 \\ \hline

    FALCON & \makecell[tl]{~Reduced\\~per-class} & 5 & 60k$\times$5 & $\sim$50~
    & 15.6 & 25.7 & 163.8 & 11.1 & 25.1 & 22.6 & 9.7 & 30.8 & 1.1 & 6.9 & 5.1 & 31.9 & 28.2 & 64.7 & 1.6 & 4.0 & 31.74 \\
    
    \bottomrule
    \end{tabular}}
\end{table*}


\subsection{Cross-Topology Knowledge Transfer}
\label{subsec:TL}

To evaluate cross-topology generalization, selected circuit topologies (RVCO, ClassEPA, DBAMixer, CSLNA, and CVA) are excluded during class training and treated as unseen targets. The model is first trained using 30\%/35\%/35\% training/validation/test splits on the three topologies within each class and a single layer is fine-tuned on the held-out topology with the same ratio (\textit{reduced per-class}), with the average number of test samples in each classes same as Figures~\ref{fig:PA}--\ref{fig:Mixer}.
\new{The same process is repeated for 70\%/15\%/15\% ratios for this work (\textit{per-class}). For FALCON, \textit{reduced per-class} (30\%/35\%/35\%), \textit{reduced} (30\%/35\%/35\%), and \textit{complete} (70\%/15\%/15\%) settings, similar to Subsection~\ref{subsec:SoTA}, are tested. The knowledge transfer performance is reported in Table~\ref{tab:FineTune}.

The average weighted MRE of this work while fine-tuning the first to last layer is 1.1\%, 1.2\%, 1.5\%, 2.3\%, and 20.5\% in the \textit{per-class} mode. 
The RF-informed indexing places every device of the held-out topology at designated functional indices; therefore, all held-out circuits can absorb the resulting small distribution shifts by fine-tuning a single nonlinear layer (i.e., GNN's). The layer at which fine-tuning is optimal, however, depends on whether the FoM behavior is covered in the initial training set. Concretely, as intuitively described in Section~\ref{sec:III}:
\begin{enumerate}[leftmargin=*]
    \item If the underlying mechanism of the FoM was present during initial training, it is the device-level sizing information that should be transferred to held-out topologies, precisely the role of the first layer. Subcircuits are commonly repeated within RF classes, leading to similar device functionalities in most circuits and FoMs. Examples include DBAMixer combining the double-balanced structure of DBPMixer with the active transconductance stage of SBAMixer, and the common-source stage of CSLNA embedded within CLNA in training set.
    \item If the underlying mechanism of the FoM was not present during initial training, fine-tuning the fourth layer yields the best results, as this layer is in charge of long functional chains. The only example includes V\textsubscript{Gain} of the CVA, which is proportional to inter-device impedance ratio, requires understanding of load and tail transistors in a cascode, and is not covered by the single-gain-path VAs in training set. Needless to say, local, single-device quantities such as P\textsubscript{DC} and BW are still best transferred through the first layer, as they are governed by local, single-device quantities.
\end{enumerate}
In practice, cases in (2) remain rare, as the GNN is trained on topologically-versatile examples to enable optimal generalization to held-out topologies through fine-tuning the first layer, which is therefore taken as the reference.

FALCON, in contrast, encodes device information through node and edge MLP encoders at the first layer before the GNN message passing, making topology-level information not pass through early. 
As a result, fine-tuning the first layers of MLP encoder and GNN simultaneously degrades performance to 560.8\% weighted MRE. Fine-tuning the final MLP head, in charge of shaping output distribution, yields the best average weighted MRE of 28.8\%.
This work outperforms these numbers by fine-tuning any GNN layer, showing effective knowledge transfer of specially gain-centric topologies (e.g., CSLNA, CVA), which due to the topology- and role-unaware MLP encoder in FALCON, produces out-of-distribution embeddings. Quantitatively, this work reduces weighted MRE on held-out circuits from 28.8\% to 1.1\%, an improvement of $\sim$26.2$\times$. Reducing the training dataset size by $\sim$2.3$\times$ leads to weighted MRE of 1.5\% in this work and 189.1\% in FALCON, highlighting data efficiency capabilities. Furthermore, similar to Subsection~\ref{subsec:SoTA}, it is concluded that forcing a unified model to be per-class with the same training dataset size worsens performance, increasing the weighted MRE from 189.1\% to 224.0\%, a degradation of $\sim$1.2$\times$.}

\subsection{Timing Results}
\label{subsec:timing}

\begin{figure}[t]
  \vspace{-3pt}
  \centering
  \includegraphics[width=\columnwidth]{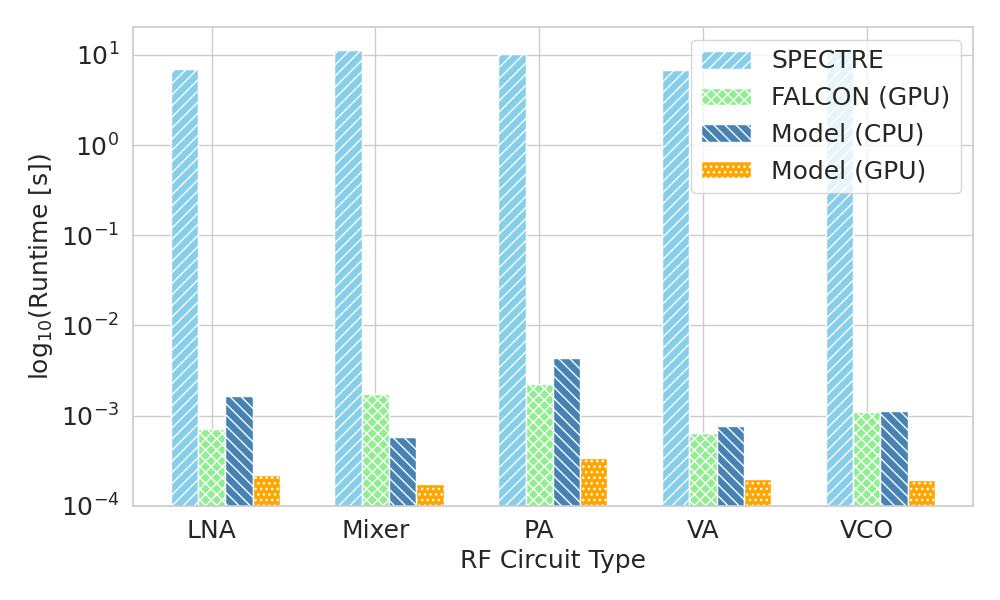}
  \vspace{-15pt}
  \caption{Comparison of per-circuit average inference latency for different RF circuits (LNA, Mixer, PA, VA, and VCO) with the model on CPU (0.57-4.37 milliseconds) and GPU (0.17-0.34 milliseconds) versus per-circuit average Spectre simulations runtime (6.73-11.69 seconds) and FALCON inference latency on GPU (0.63-2.22 milliseconds).} 
  \label{fig:TimingAnalysis}
\end{figure}

Test or inference latency is measured as average per-sample inference time under batched evaluation, shown in Figure~\ref{fig:TimingAnalysis}. The reported latency includes CPU-GPU transfer of precomputed graph data and model inference latency but, similar to training time, excludes offline preprocessing steps such as netlist parsing and graph construction. Compared to an average Spectre simulation runtime of 9.384~s per circuit, the proposed surrogate reduces evaluation time to 1.687~ms on CPU and 0.225~ms on GPU, corresponding to speedups of approximately $5.6\times10^{3}$ and $4.2\times10^{4}$, respectively. 

\section{Conclusion and Future Work}
\label{sec:VI}

A class-specialized GNN model is presented for accurate and data-efficient prediction of a variety of RFICs and performance metrics. The transistor-level framework proposes functionality-aware feature indexing that achieves an average test MRE of 2.71\% across nineteen topologies, an improvement by 3.3$\times$ in prior art. To reach the MRE of 9.09\% reported by prior art \cite{g19}, this work requires $\sim$36$\times$ less training data. Furthermore, compared to previous works, knowledge transfer to held-out topologies is improved by $\sim$26.2$\times$ through lightweight fine-tuning a single GNN layer, without overhead of an extra encoder.

A class-specialized model is favorable due to its fast training time and natural alignment with practical RF design workflows that search within a desired class. However, forcing a unified model to be class-specialized deteriorates accuracy by $\sim$2.3$\times$ and cross-topology generalization by $\sim$1.2$\times$. Building on this, this work demonstrates $\sim$20$\times$ faster training, demonstrating its effectiveness for scalable and deployment-ready RF design.

\new{While this work targets schematic-level surrogate modeling, it supports post-layout area optimization of passive structures through existing frameworks (e.g., \cite{g19}). A fully post-layout model incorporating interconnect parasitics (capacitive, resistive, and inductive), routing effects, and full post-layout behavior remains an important direction for future work.}

\section{Acknowledgment}

\new{The authors used ChatGPT solely to improve the readability and language of the manuscript. All generated text was carefully reviewed and edited by the authors, who take full responsibility for the content and accuracy of this work.}

\bibliographystyle{IEEEtran}
\bibliography{references}

\vfill\eject

\begin{IEEEbiography}[{\includegraphics[width=1in,height=1.25in,clip,keepaspectratio]{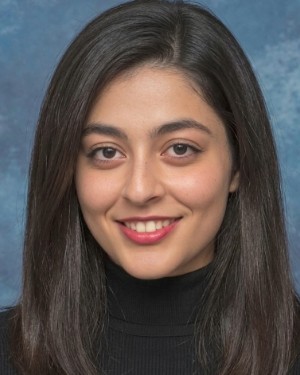}}]{Anahita Asadi}
(Graduate Student Member, IEEE) received the B.Sc. degree in electrical engineering from K. N. Toosi University of Technology, Tehran, Iran, in 2022. She is currently pursuing the Ph.D. degree in electrical and computer engineering at the University of Illinois Chicago, Chicago, IL, USA.

Since 2023, she has been a Graduate Research Assistant with the University of Illinois Chicago, where she has developed graph neural network and active learning frameworks for RF circuit performance prediction and optimization. In 2026, she joined Keysight Technologies, Santa Rosa, CA, USA, as a Physical AI Research Intern. Her research interests include machine learning on graphs, physics-informed neural networks, and AI surrogate models for analog and RF circuit design and optimization.

She is a recipient of the IEEE/ACM Design Automation Conference (DAC) Young Fellowship in 2025 and 2026.
\end{IEEEbiography}

\begin{IEEEbiography}[{\includegraphics[width=1in,height=1.25in,clip,keepaspectratio]{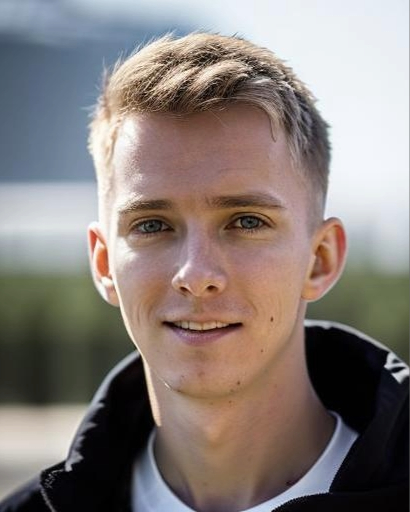}}]{Leonid Popryho}
(Graduate Student Member, IEEE) received the B.Sc. degree in applied mathematics from Igor Sikorsky Kyiv Polytechnic Institute, Kyiv, Ukraine, in 2021, and the dual M.Sc. degree in computer science from Blekinge Institute of Technology, Sweden, and Kyiv Academic University, Ukraine, in 2023. He is currently pursuing the Ph.D. degree in electrical and computer engineering at the University of Illinois Chicago, IL, USA.

Since 2023, he has been a Graduate Research Assistant with the HiPerCAS Laboratory, University of Illinois Chicago, and in 2026 joined the X-ray Science Division, Argonne National Laboratory, Lemont, IL, USA, as a Research Aide. His research interests include machine learning for electronic design automation, graph neural networks, physics-informed neural networks, active learning, and surrogate modeling for device and circuit optimization.

He was a recipient of the IEEE/ACM Design Automation Conference (DAC) Young Fellowship in 2025 and 2026.
\end{IEEEbiography}

\begin{IEEEbiography}[{\includegraphics[width=1in,height=1.25in,clip,keepaspectratio]{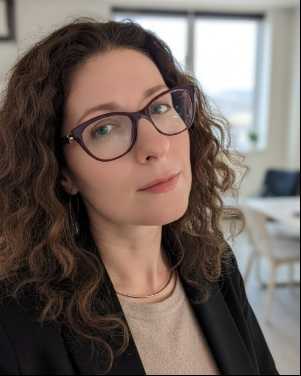}}]{Inna Partin-Vaisband}
(Senior Member, IEEE) received the Ph.D. degree in electrical and computer engineering from the University of Rochester, Rochester, NY, USA, in 2015, and the B.Sc. and M.Sc. degrees from the Technion-Israel Institute of Technology, Haifa, Israel, in 2006 and 2009, respectively.

She was with IBM from 2005 to 2009. She is currently an Associate Professor of electrical and computer engineering at the University of Illinois Chicago, Chicago, IL, USA. Her research focuses on enabling 2.5D/3D high-performance computing platforms through co-design methodologies, automation, and power delivery.

Dr. Partin-Vaisband serves as an Associate Editor for the IEEE Transactions on Components, Packaging and Manufacturing Technology and the IEEE Circuits and Systems Magazine. She was a recipient of the 2022 Google Research Scholar Award and the 2023 NSF CAREER Award.
\end{IEEEbiography}

\end{document}